\documentclass[10pt,twocolumn]{article}

\PassOptionsToPackage{hidelinks}{hyperref}
\usepackage{amsmath,amssymb,amsthm}
\usepackage{booktabs}
\usepackage{multirow}
\usepackage{algorithm}
\usepackage{algpseudocode}
\usepackage{graphicx}
\usepackage{authblk}
\usepackage[round,authoryear]{natbib}
\usepackage{orcidlink}
\usepackage{mathtools}
\usepackage{placeins}
\usepackage{threeparttable}

\usepackage{marvosym}
\usepackage{enumitem}
\usepackage[margin=1in]{geometry}
\usepackage[ugaramond]{mathdesign}

\usepackage{fancyhdr}
\usepackage{lastpage}
\usepackage{capt-of}
\graphicspath{{./assets/images/}{./assets/figures/}{./}}

\theoremstyle{plain}

\newtheorem{lemma}{Lemma}
\newtheorem{corollary}{Corollary}
\newtheorem{proposition}{Proposition}

\theoremstyle{definition}
\newtheorem{assumption}{Assumption}
\newtheorem{problem}{Problem}

\theoremstyle{remark}
\newtheorem{remark}{Remark}

\newcommand{\R}{\mathbb{R}}
\newcommand{\St}{\mathrm{St}}
\newcommand{\Meas}{\mathsf{H}}
\newcommand{\tr}{\operatorname{tr}}
\newcommand{\diag}{\operatorname{diag}}

\newcommand{\sym}{\operatorname{sym}}
\newcommand{\qf}{\operatorname{qf}}
\newcommand{\Retr}{\operatorname{Retr}}
\newcommand{\Tfit}{\mathcal{T}_{\mathrm{fit}}}
\newcommand{\Tsearch}{\mathcal{T}_{\mathrm{search}}}
\newcommand{\Tselect}{\mathcal{T}_{\mathrm{select}}}
\newcommand{\Ttest}{\mathcal{T}_{\mathrm{test}}}
\newcommand{\Gain}{\mathcal{G}}
\newcommand{\loss}{\ell}
\newcommand{\E}{\mathbb{E}}

\newcommand{\PaperTitle}{REGAIN: REconciliation GAIN-driven Auxiliary Direction Learning}

\newcommand{\PaperAuthorOne}{Weijia Li}
\newcommand{\PaperAuthorTwo}{Shun Hu}
\newcommand{\PaperAuthorThree}{Yanfei Kang}
\newcommand{\PaperAuthorThreeOrcid}{0000-0001-8769-6650}

\newcommand{\PaperAffiliationOne}{School of Mathematical Sciences, Beihang University, Beijing, China}
\newcommand{\PaperAffiliationTwo}{School of Economics and Management, Beihang University, Beijing, China}
\newcommand{\PaperCorrespondingEmail}{yanfeikang@buaa.edu.cn}

\newcommand{\PaperCorrespondence}{
Corresponding author: Yanfei Kang (\texttt{\PaperCorrespondingEmail})
}
\newcommand{\PaperKeywords}{forecast reconciliation | auxiliary direction learning | linear component augmentation | hierarchical forecasting}
\newcommand{\PaperAbstractText}{Forecast reconciliation usually starts from a fixed measurement system and asks
how forecasts should be projected onto a coherent space. We ask a different
question: which additional linear measurements should be forecast and included
in the reconciliation system? We propose \textsc{REGAIN}, a reconciliation-gain
framework that learns normalized auxiliary directions, forecasts the induced
series with a frozen forecasting oracle, and selects directions by their
target-weighted loss reduction after augmented generalized least-squares
reconciliation. Unlike variance-based components or predictability-based
auxiliary selection, REGAIN optimizes the downstream effect of an auxiliary
measurement on the final reconciled forecasts. 
We provide a statistical characterization showing that useful auxiliary directions
must provide complementary information about unresolved target uncertainty, 
rather than merely being easy to forecast. The analysis also clarifies the covariance-risk reduction mechanism, 
the role of bias changes in realized quadratic risk, and the stability of estimated gain signals. 
A stagewise learning algorithm with held-out gain screening is developed, together with an optional joint refinement step. Experiments on Beijing PM2.5 and Australian Tourism data show that gain-selected measurements can improve both ordinary
multivariate and hierarchical forecasts, especially when they reveal residual
uncertainty not captured by the original measurement system.}

\begin{document}

\title{\PaperTitle}

\author[1]{\PaperAuthorOne}
\author[2]{\PaperAuthorTwo}
\author[2]{\PaperAuthorThree\orcidlink{\PaperAuthorThreeOrcid}\thanks{\PaperCorrespondence}}

\affil[1]{\PaperAffiliationOne}
\affil[2]{\PaperAffiliationTwo}

\date{}

\maketitle

\begin{abstract}
\PaperAbstractText
\end{abstract}

\noindent\textbf{Keywords:} \PaperKeywords
\vspace{1em}

\section{Introduction}

Forecast reconciliation is usually studied after the measurement system has already been specified. In hierarchical and grouped time series, a state vector \(b_t\), often representing bottom-level series, is linked to the reported series through a known measurement matrix \(S\), and the standard task is to transform base forecasts into forecasts that are coherent with that structure. Classical reconciliation methods derive such transformations from generalized least-squares arguments, while more recent work learns improved projection rules from data \citep{Hyndman2011OptimalCombination,Wickramasuriya2019MinT,Panagiotelis2023ProbabilisticReconciliation,Tsiourvas2024LearningOptimalProjection}. A related line of work shows that the structure itself can matter: temporal, cross-temporal, clustered, or otherwise constructed hierarchies may change the effectiveness of reconciliation \citep{Athanasopoulos2017TemporalHierarchies,DiFonzo2023CrossTemporal,Zhang2025Constructing}. Together, these lines establish a common perspective: the quality of reconciled forecasts depends both on how one reconciles and on what measurement structure is available. This naturally raises a more general question: can we learn additional linear measurements that improve the final target-node forecasts?

This question is not limited to strict hierarchical settings. In an ordinary multivariate forecasting problem, the natural measurement matrix is simply the identity, yet one may still form auxiliary linear series from the observed variables and reconcile them jointly with the original forecasts. Recent linear component augmentation work shows that such additional components can reduce forecast error variance when they are available for joint projection \citep{Yang2024FLAP}. Component-analysis and dimension-reduction methods also suggest that useful statistical structure need not align with the raw coordinates; different transformations may target variance, predictability, decorrelation, or lower-dimensional dependence \citep{Goerg2013ForeCA,Matteson2011DOC,Banerjee2020FPCAChangePoint,Liang2020NDRPerformance,Zhu2022DistributedDR,Li2022LocalSVMDR,Liang2024IQVAR}. 

Yet these approaches do not directly optimize the downstream reconciliation objective. Components selected by variance explained or standalone predictability may add little to the final reconciled forecast if their forecast errors are redundant with those of the original series. Conversely, a less predictable component may still be valuable if it carries complementary information about target-relevant uncertainty. Therefore, the goal is not merely to find a forecastable component, but to learn an auxiliary measurement whose inclusion improves final reconciled risk.

This paper addresses this missing layer by learning auxiliary directions directly from final reconciliation gain. While existing approaches optimize the reconciliation map, construct alternative hierarchy structures, or evaluate pre-defined component families such as principal components, our approach treats the auxiliary directions themselves as learnable objects. The key idea is that an auxiliary direction should be judged by the gain it delivers after being forecast and integrated into the augmented measurement system.

Grounded in this perspective, we propose \textsc{REGAIN}---\emph{REconciliation GAIN-driven Auxiliary Direction Learning}---as a framework for learning such measurements. Given a fixed natural measurement matrix \(S\), we introduce a small matrix of auxiliary directions \(U\) and form auxiliary series \(c_t(U)=U^\top b_t\). Rather than recursively retraining predictors, we utilize frozen time-series foundation models (TSFMs)~\citep{Das2024TimesFM,Ansari2024Chronos} as a shared forecasting oracle. The induced auxiliary forecasts are stacked with the original base forecasts and reconciled via an augmented GLS system. Candidate directions are systematically evaluated by their empirical direct gain: the target-weighted loss reduction of the final reconciled forecasts relative to standard, no-auxiliary reconciliation. This frozen-oracle design helps isolate the effect of measurement design from artifacts of forecaster adaptation.

Our contributions are threefold. First, we formulate auxiliary measurement design for forecast reconciliation as a downstream direction-learning problem: auxiliary series are selected not by variance explained or standalone forecastability, but by the target-weighted loss reduction they produce after augmented reconciliation. Second, we provide a gain-based statistical characterization of auxiliary usefulness. Under correctly specified population GLS, augmentation satisfies a covariance-risk non-deterioration property, while realized quadratic-risk or test-loss improvement can still depend on bias changes; the single-direction formula identifies the roles of target exposure, complementary residual information, and effective auxiliary noise. Third, we develop a stagewise learning procedure with held-out marginal-gain screening and an optional joint refinement step. Experiments on multivariate and hierarchical benchmarks show that gain-selected directions can improve final target-node accuracy, with the strongest gains arising when the auxiliary directions capture residual uncertainty not already resolved by the natural forecast block.

\begin{figure*}[t]
\centering
\includegraphics[width=0.98\linewidth]{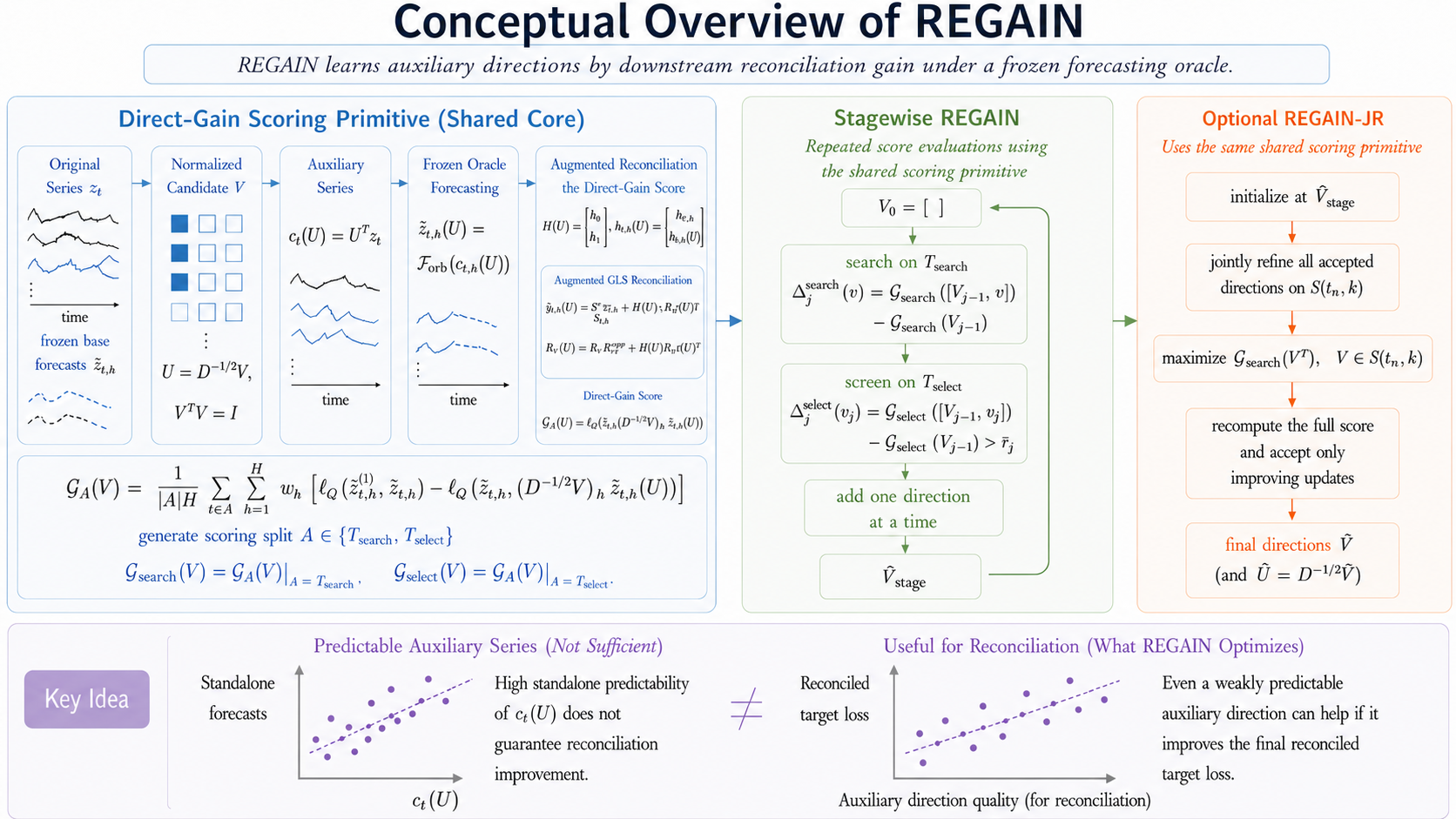}
\caption{
Conceptual overview of REGAIN. Stagewise REGAIN is the core procedure for
learning auxiliary directions by downstream reconciliation gain, while
REGAIN-JR is an optional refinement step that further adjusts the accepted
directions.
}
\label{fig:method-overview}
\end{figure*}

\section{Related Work}
\label{sec:related-work}

Forecast reconciliation can be viewed as using additional measurement relations to correct forecast errors in a set of target series. Most existing studies take these relations as given and focus on how to project incoherent forecasts back to the coherent space. A smaller but growing literature changes the measurement system itself, for example through temporal or cross-temporal aggregation, constructed hierarchies, or auxiliary components. Recent foundation models add another useful ingredient: they make it possible to evaluate many candidate auxiliary series under a common forecasting oracle. These developments motivate the question studied here: rather than choosing auxiliary series by interpretability, variance explained, or standalone forecastability, can we learn the linear directions that most improve the final reconciled forecasts? We review the literature from this perspective and position \textsc{REGAIN} as a gain-driven approach to auxiliary measurement design.

\subsection{Forecast reconciliation with fixed measurement systems}

Classical forecast reconciliation starts from a known aggregation or measurement matrix and asks how incoherent base forecasts should be projected back to the coherent space \citep{Hyndman2011OptimalCombination,Wickramasuriya2019MinT}. The key object is the reconciliation map, typically derived from generalized least-squares arguments and from an estimate of the base-forecast error covariance matrix. Probabilistic reconciliation extends the same principle from point forecasts to predictive distributions, while preserving the assumption that the measurement system is already given \citep{Panagiotelis2023ProbabilisticReconciliation,Girolimetto2024CrossTemporalProb}. Recent learning-based reconciliation methods further replace analytically specified projections with data-adaptive projection rules, but still operate within a supplied hierarchical or grouped structure \citep{Tsiourvas2024LearningOptimalProjection}.

This literature provides the statistical foundation for our augmented projection step. The distinction is that its optimization variable is usually the map used to reconcile forecasts, not the measurement rows available to the map. In contrast, we take the natural structure as the baseline and ask whether additional rows, represented by learned directions, should be introduced before reconciliation. Thus the role of GLS in our framework is not only to enforce coherence under a fixed structure, but also to evaluate the downstream value of candidate auxiliary measurements.

\subsection{Structure construction and alternative hierarchies}

A second line of work shows that the measurement structure itself can affect forecast accuracy. Temporal and cross-temporal reconciliation add aggregation constraints across forecast horizons as well as across series, thereby changing the space in which coherence is imposed \citep{Athanasopoulos2017TemporalHierarchies,DiFonzo2023CrossTemporal,Girolimetto2024CrossTemporalProb}. Other work constructs hierarchies from the data, compares random or permuted hierarchies with the same shape, or averages forecasts across multiple candidate hierarchies \citep{Zhang2025Constructing}. The common message is that reconciliation performance depends not only on the reconciliation algorithm, but also on which measurement relations are made available.

Our work shares this message, but the intervention is different. Constructed-hierarchy approaches usually search over interpretable aggregation structures or over alternative hierarchical partitions. \textsc{REGAIN} does not replace the supplied hierarchy, nor does it require the learned directions to be valid aggregate nodes. Instead, it appends a small number of auxiliary linear measurements outside the natural structure and judges them only by their contribution to final target-node risk. This makes the method applicable both when a hierarchy is present and when the natural measurement system is simply the identity matrix.

\subsection{Linear component augmentation for multivariate forecasting}

The closest related work studies linear component augmentation for multivariate forecasting. FLAP shows that augmenting the original series with additional linear combinations and then projecting jointly can reduce forecast-error variance, with principal components serving as a natural default component family \citep{Yang2024FLAP}. Earlier component-analysis methods also construct transformed series, but typically optimize criteria such as predictability, decorrelation, or dynamic variation rather than final reconciliation gain \citep{Goerg2013ForeCA,Matteson2011DOC}. More broadly, nonlinear dimension-reduction assessment, sufficient-dimension-reduction methods, and robust multivariate spatio-temporal modeling all support the view that useful statistical structure need not align with the raw coordinates \citep{Liang2020NDRPerformance,Banerjee2020FPCAChangePoint,Zhu2022DistributedDR,Li2022LocalSVMDR,Liang2024IQVAR}. Recent neural multivariate forecasting architectures make a similar point through learnable decomposition, dual-attention designs, and adaptive basis construction \citep{Yu2024Leddam,Ni2023BasisFormer}.

\textsc{REGAIN} adopts the same broad premise that transformed linear measurements can be useful, but it differs in the criterion used to choose them. A component that is highly variable or easy to forecast may be nearly redundant with the original forecast errors and therefore contribute little after reconciliation. Conversely, a component with only moderate standalone predictability may still be valuable if its forecast error carries non-redundant information about unresolved target uncertainty. We therefore treat the auxiliary direction matrix as a decision variable and score each candidate by the reduction it produces in final reconciled loss. This moves the problem from component discovery for representation quality or base-forecast accuracy to direction learning for downstream reconciliation gain.

\subsection{Foundation-model forecasting as a frozen oracle}

Time-series foundation models and other pre-trained or zero-shot forecasters
provide a practical way to forecast many natural and auxiliary series under a
common forecasting mechanism
\citep{Das2024TimesFM,Ansari2024Chronos,Zhou2023OneFitsAll,Ekambaram2024TTM,Wang2025UnifiedRepresentation,Dooley2023ForecastPFN,Gruver2023LLMTime,Wang2024TimeXer}.
In our framework, they are used only as frozen shared oracles, not as the
theoretical object of study. The same oracle, forecast mode, and cached base
forecasts are held fixed across candidate directions and baselines. This design
makes comparisons fair and isolates the effect of the learned measurement
directions: a direction is useful only if, after being forecast by the same
oracle and combined through the same augmented GLS rule, it reduces the final
target-weighted reconciliation loss.

\section{Problem Setup}
\label{sec:problem-setup}

Let \(b_t\in\R^n\) denote the state to be reconciled and let
\(S\in\R^{m\times n}\) be a fixed natural measurement matrix. The
reported or evaluated series are
\[
y_t = S b_t .
\]
This notation covers both settings used in the paper. In an ordinary
multivariate problem, \(S=I_n\) and \(y_t=b_t\). In a hierarchical or
grouped problem, \(b_t\) is the bottom-level vector and \(S\) is the
supplied aggregation matrix.

For each rolling origin \(t\) and horizon \(h\in\{1,\dots,H\}\), a frozen
forecasting model produces a base forecast
\[
\hat y_{t,h}\in\R^m
\]
for the natural measurements. REGAIN seeks to improve the final
reconciled forecasts by adding a small number of auxiliary linear
measurements of the state. Let
\[
U=[u_1,\dots,u_k]\in\R^{n\times k}
\]
collect the auxiliary directions, and define
\[
c_t(U)=U^\top b_t\in\R^k.
\]
The same frozen forecasting model is applied to the historical auxiliary
series to obtain direct forecasts
\[
\hat c_{t,h}(U)\in\R^k.
\]
More explicitly, let \(\mathcal F_{\theta,h}\) denote the \(h\)-step
forecasting operator with parameter vector \(\theta\), and write
\[
\hat c_{t,h}(U)
=
\mathcal F_{\theta,h}\big(c_{1:t}(U)\big),
\qquad
c_{1:t}(U)=(U^\top b_1,\dots,U^\top b_t).
\]
Throughout this formulation, \(\theta\) is frozen and the optimization
variable is only the auxiliary-direction matrix \(U\), or its whitened
counterpart \(V\) below. Thus REGAIN does not retrain the predictor for
each candidate direction; it treats the predictor as a fixed map from the
input auxiliary history to the forecast it returns.

The required oracle regularity depends on the claim being made. For the
fixed-\(U\) definitions and the population GLS risk comparisons, it is enough
that the frozen oracle returns forecasts for each candidate \(U\). The
compactness argument for the normalized search problem additionally requires
the resulting direct-gain map to be continuous in \(U\). The gradient-based JR
implementation uses differentiability of the local surrogate with respect to
the normalized coordinates \(V\); if a frozen oracle is available only as a
black-box or nonsmooth map, the same gain objective can instead be optimized by
a derivative-free local search.

For a fixed \(U\), define the augmented measurement matrix and the
augmented forecast vector as
\[
\Meas_S(U)=
\begin{bmatrix}
S\\
U^\top
\end{bmatrix},
\qquad
\hat z^S_{t,h}(U)=
\begin{bmatrix}
\hat y_{t,h}\\
\hat c_{t,h}(U)
\end{bmatrix}.
\]
Here
\[
e^y_{t,h}=y_{t+h}-\hat y_{t,h},
\qquad
e^{c,S}_{t,h}(U)=U^\top b_{t+h}-\hat c_{t,h}(U)
\]
denote the natural-measurement and auxiliary forecast errors. Given a
positive definite estimate \(W^S_h(U)\) of the joint residual covariance
of the stacked forecast errors
\((e^y_{t,h},e^{c,S}_{t,h}(U))\), the augmented state forecast is defined
by the regularized generalized least-squares problem
\begin{equation}
\label{eq:augmented-gls}
\begin{aligned}
\tilde b_{t,h}(U)
\in
\arg\min_{b\in\R^n}\quad
&\big(\hat z^S_{t,h}(U)-\Meas_S(U)b\big)^\top
\big(W^S_h(U)\big)^{-1}\\
&\big(\hat z^S_{t,h}(U)-\Meas_S(U)b\big)
+\eta_{\mathrm{gls}}\|b\|_2^2,
\end{aligned}
\end{equation}
where \(\eta_{\mathrm{gls}}\geq 0\) is a small GLS stabilization parameter.
The theoretical population statements below set \(\eta_{\mathrm{gls}}=0\), while
the finite-sample implementation uses a small scale-adaptive ridge. The reconciled
forecast on the reported natural measurements is
\[
\tilde y_{t,h}(U)=S\tilde b_{t,h}(U).
\]

Let \(\tilde b^{(0)}_{t,h}\) and
\(\tilde y^{(0)}_{t,h}=S\tilde b^{(0)}_{t,h}\) denote the natural-only
reconciled forecasts obtained from the measurement block \(S\) alone.
The reported-node reconciliation errors are
\[
\varepsilon^y_{0,t,h}=\tilde y^{(0)}_{t,h}-y_{t+h},
\qquad
\varepsilon^y_{U,t,h}=\tilde y_{t,h}(U)-y_{t+h}.
\]
Let \(\mathcal T_{\mathrm{search}}\) denote the rolling-origin split used
to score candidate directions during search. For any positive
semidefinite target weight matrix \(Q\) on the reported variables, write
\[
\begin{aligned}
\loss_Q(a,b)
&=
(a-b)^\top Q(a-b).
\end{aligned}
\]
The empirical direct gain of \(U\) is
\begin{equation}
\label{eq:empirical-direct-gain}
\begin{aligned}
\mathcal G_S(U)
&=
\frac{1}{|\mathcal T_{\mathrm{search}}|}
\sum_{t\in\mathcal T_{\mathrm{search}}}
\sum_{h=1}^H
\omega_h
\left[
\loss_Q(\tilde y^{(0)}_{t,h},y_{t+h})
\right.\\
&\qquad\left.
-
\loss_Q(\tilde y_{t,h}(U),y_{t+h})
\right],
\end{aligned}
\end{equation}
where \(\omega_h\ge 0\) and \(\sum_{h=1}^H\omega_h=1\). When the
measurement matrix is clear from context, we write this direct-gain
functional simply as \(\Gain(U)\).

A natural but incomplete alternative is to rank directions by the standalone forecastability of the induced auxiliary series, for example by minimizing
\[
\mathcal P(U)
=
\frac{1}{|\mathcal T_{\mathrm{search}}|}
\sum_{t\in\mathcal T_{\mathrm{search}}}
\sum_{h=1}^H
\omega_h
\left\|
\hat c_{t,h}(U)-U^\top b_{t+h}
\right\|_2^2 .
\]
This proxy measures the auxiliary forecast loss alone, but not whether the auxiliary forecast errors provide non-redundant information that reduces the reconciled risk on the target variables. The search problem is therefore to maximize the downstream gain in Equation~\eqref{eq:empirical-direct-gain}, written as $\Gain(U)$ when the measurement matrix is clear, rather than to minimize $\mathcal P(U)$ alone.

The direct objective should not be optimized over an unconstrained matrix \(U\), because the direction matrix $U$ contains parameterization redundancy. Rescaling a column changes the scale of the auxiliary series presented to the frozen forecaster, while repeated or nearly collinear columns spend degrees of freedom on the same effective direction and tend to create unstable search behavior. To remove this redundancy, we introduce a positive definite scaling matrix $D \succ 0$ and impose the normalization
\[
U^\top D U = I_k.
\]
Here \(D\) is a scaling matrix used to define scale-free directions; it is not
a covariance model for the state, the auxiliary series, or the forecast errors.
This constraint simultaneously fixes the scale of each auxiliary direction relative to $D$ and prevents the collection from collapsing onto duplicated directions.
The orthogonality requirement should therefore be read as a normalization of
the search geometry rather than as a statistical assumption that useful
auxiliary measurements, or their forecast errors, must be independent.
Intermediate angular relations between candidate directions are of course
possible, but allowing nearly collinear columns would let several auxiliary
coordinates spend capacity on almost the same measurement. In that case the
observed gain would be harder to attribute to genuinely new directional
information rather than to rescaling, mixing, or duplication of directions.
The statistical dependence that matters for reconciliation is still retained
in the covariance blocks \(R^S_h(U)\) and \(K^S_h(U)\) used by the augmented
GLS projection.

With the whitening change of variables
\[
V = D^{1/2}U,
\]
the constraint becomes
\[
V^\top V = I_k.
\]
Hence the feasible set is
\[
\St(n,k)=\{V \in \R^{n \times k}: V^\top V = I_k\},
\]
\begin{problem}[Normalized Stiefel direct-gain formulation]
\label{prob:normalized-gain}
Given a fixed natural measurement matrix \(S\), a positive-definite
scaling matrix \(D\), an auxiliary dimension \(1\le k\le n\), and the
fixed-\(U\) augmented forecast defined by Equation~\eqref{eq:augmented-gls}, find
\begin{equation}
\label{eq:normalized-stiefel-objective}
V^\star
\in
\arg\max_{V\in \St(n,k)}
\mathcal G_S(D^{-1/2}V).
\end{equation}
\end{problem}
This Stiefel formulation is not merely a numerical constraint: it defines
the normalized auxiliary-direction class over which direct gain is compared.
The distinction between Stiefel and Grassmann geometry is standard in
orthogonality-constrained optimization \citep{Edelman1998Geometry,Absil2008Optimization};
here it matters because each column of \(U\) generates a coordinate
auxiliary series, and the frozen oracle need not be invariant under a
basis rotation \(U\mapsto UQ\).

Since \(\St(n,k)\) is compact, the normalized problem is well posed
whenever the direct-gain map is continuous; a compactness argument is
provided in Appendix~\ref{app:stiefel-wellposedness}.

\section{Direct-Gain Characterization}
\label{sec:direct-gain-characterization}

Section~\ref{sec:problem-setup} defines empirical direct gain as the
criterion for evaluating candidate auxiliary directions. This section gives
the population interpretation of that criterion and separates three related
objects that must not be conflated. The first is the direct quadratic-risk
gain, which is the population counterpart of the loss reduction optimized by
\textsc{REGAIN}. The second is an analytical covariance-risk gain, which
isolates the variance-side mechanism of augmented GLS under correctly
specified covariances. The third is the finite-sample gain score used by the
algorithm, where covariance matrices are estimated and candidate directions
are evaluated on held-out rolling origins.

The distinction matters because an auxiliary direction can help only through
its effect after forecasting and reconciliation. Standalone predictability
does not determine this effect, and neither does a covariance calculation by
itself once bias and finite-sample estimation are present.

Throughout this section, we fix a forecast horizon \(h\) and keep the
natural measurement matrix \(S\) explicit. The ordinary multivariate case is
obtained by setting \(S=I_n\), so that \(y_t=b_t\). For the theoretical
statements we allow a horizon-specific target weight \(Q_h\succeq 0\); the
empirical objective in Equation~\eqref{eq:empirical-direct-gain} is the
special case \(Q_h\equiv Q\), followed by the horizon average with weights
\(\omega_h\). The population formulas below use the unregularized GLS
projection to expose the statistical structure. The ridge and shrinkage
terms used by the implemented method are finite-sample stabilizers and are
introduced in Section~\ref{sec:direct-gain-scoring}.

The assumptions and main statements are given here, and the algebraic proofs
are deferred to Appendix~A, which restates only the proof-relevant pieces for
convenience.

Define the residuals for the natural measurements and the auxiliary
series by
\[
e^y_{t,h}=y_{t+h}-\hat y_{t,h},
\qquad
e^{c,S}_{t,h}(U)=U^\top b_{t+h}-\hat c_{t,h}(U),
\]
and write their joint residual covariance as
\[
W^S_h(U)=
\begin{bmatrix}
		W_{yy,h} & K^S_h(U) \\
		K^S_h(U)^\top & R^S_h(U)
\end{bmatrix}.
\]
Here \(W_{yy,h}\) is the covariance of natural-measurement residuals,
\(R^S_h(U)\) is the covariance of auxiliary residuals, and \(K^S_h(U)\)
is the cross-covariance between the two blocks.
\begin{assumption}[Second moments and well-posed GLS systems]
\label{ass:wellposed}
For each horizon \(h\) and each candidate direction matrix \(U\)
considered below, the residual vectors
\(e^y_{t,h}\) and \(e^{c,S}_{t,h}(U)\) have finite second moments, so
\(W^S_h(U)\) and its blocks are well defined. We assume that
\(W_{yy,h}\) and \(W^S_h(U)\) are positive definite, and that the
bottom-level normal matrices
\[
S^\top W^{-1}_{yy,h}S
\quad\text{and}\quad
\Meas_S(U)^\top[W^S_h(U)]^{-1}\Meas_S(U)
\]
are invertible.
\end{assumption}

Assumption~\ref{ass:wellposed} ensures that the covariance objects used
below are well defined and allows the natural-only and augmented
reconciliation formulas to be written in their unregularized GLS form.
The small ridge and shrinkage terms used in computation are numerical
safeguards and do not change the conceptual role of the population
formulas.

\subsection{Risk interpretation of the search gain}
\label{sec:risk-interpretation-search-gain}

Throughout the population statements in this subsection,
\(\tilde b^{(0)}_{t,h}\) and \(\tilde b_{t,h}(U)\) denote the population
GLS analogues of the
natural-only and augmented reconciled state forecasts: the ridge term in
Equation~\eqref{eq:augmented-gls} is set to zero, and the GLS weights are
the population residual covariances \(W_{yy,h}\) and \(W^S_h(U)\). Their
reported forecasts are
\[
\tilde y^{(0)}_{t,h}=S\tilde b^{(0)}_{t,h},
\qquad
\tilde y_{t,h}(U)=S\tilde b_{t,h}(U).
\]
Define the corresponding state and reported forecast errors by
\[
\varepsilon^S_{0,t,h}=\tilde b^{(0)}_{t,h}-b_{t+h},
\qquad
\varepsilon^S_{U,t,h}=\tilde b_{t,h}(U)-b_{t+h},
\]
\[
\varepsilon^y_{0,t,h}=S\varepsilon^S_{0,t,h},
\qquad
\varepsilon^y_{U,t,h}=S\varepsilon^S_{U,t,h}.
\]
These definitions are compatible with the reported-error notation in
Section~\ref{sec:problem-setup}; here they are tied specifically to the
population GLS objects used in the theoretical statements.

For a fixed horizon \(h\), the population target-weighted quadratic-risk
gain is
\begin{equation}
\label{eq:population-quad-gain}
\mathcal G^{\mathrm{quad}}_{h,S}(U;Q_h)
=
\mathbb E\!\left[
\loss_{Q_h}(\tilde y^{(0)}_{t,h},y_{t+h})
-
\loss_{Q_h}(\tilde y_{t,h}(U),y_{t+h})
\right].
\end{equation}
The expectation is taken under the population law of the forecast and
realization pair. The empirical search score used by REGAIN estimates this
same downstream risk contrast directly, after replacing the population GLS
objects by the implemented reconciled forecasts and averaging over
rolling-origin evaluations. For a single horizon this gives
\begin{equation}
\label{eq:horizon-sample-quad-gain}
\begin{aligned}
\widehat{\mathcal G}^{\mathrm{quad}}_{h,S}(U;Q_h)
&=
\frac{1}{|\mathcal T_{\mathrm{search}}|}
\sum_{t\in\mathcal T_{\mathrm{search}}}
\left[
\loss_{Q_h}(\tilde y^{(0)}_{t,h},y_{t+h})
\right.\\
&\qquad\left.
-
\loss_{Q_h}(\tilde y_{t,h}(U),y_{t+h})
\right].
\end{aligned}
\end{equation}
The multi-horizon direct gain averages these horizon-specific sample gains
with the weights \(\omega_h\). When the same target weight \(Q\) is used at
all horizons, this reduces to Equation~\eqref{eq:empirical-direct-gain}.

To understand why such an improvement can arise, we first isolate the
covariance-side component of this risk improvement. For \(j\in\{0,U\}\),
let
\[
\mathcal R^{\mathrm{cov}}_{j,h}(Q_h)
=
\operatorname{tr}\!\left(
Q_h\operatorname{Cov}(\varepsilon^y_{j,t,h})
\right)
\]
denote the target-weighted covariance risk of the reported error. For any
\(Q_h\succeq 0\), define the analytical covariance-risk gain at horizon
\(h\) as
\begin{equation}
\label{eq:analytical-cov-risk-definition}
\mathcal G^{\mathrm{ana}}_{h,S}(U;Q_h)
=
\mathcal R^{\mathrm{cov}}_{0,h}(Q_h)
-
\mathcal R^{\mathrm{cov}}_{U,h}(Q_h).
\end{equation}
The quantity in Equation~\eqref{eq:analytical-cov-risk-definition} is not
the selection objective used by REGAIN. It isolates the variance-side effect
of adding auxiliary measurements under correctly specified GLS by comparing
the target-weighted error covariance before and after augmentation.
\begin{lemma}[Baseline and augmented covariance]
\label{lem:covariance}
For the unregularized population GLS problem, under
Assumption~\ref{ass:wellposed}, the natural-only and augmented state error
covariances of the population GLS projections satisfy
\[
\begin{aligned}
\operatorname{Cov}(\varepsilon^S_{0,t,h})
&=
\Sigma^S_{0,h}
=
\big(S^\top W_{yy,h}^{-1}S\big)^{-1},\\
\operatorname{Cov}(\varepsilon^S_{U,t,h})
&=
\Sigma^S_{U,h}= \bigl(
\Meas_S(U)^\top
(W^S_h(U))^{-1}
\Meas_S(U)
\bigr)^{-1}.
\end{aligned}
\]
Consequently, the reported-error covariances are
\[
\begin{aligned}
\operatorname{Cov}(\varepsilon^y_{0,t,h})
&=
\Sigma^y_{0,h}
=
S\Sigma^S_{0,h}S^\top,\\
\operatorname{Cov}(\varepsilon^y_{U,t,h})
&=
\Sigma^y_{U,h}
=
S\Sigma^S_{U,h}S^\top.
\end{aligned}
\]
\end{lemma}

This lemma identifies the covariance object modified by auxiliary
measurements. Augmentation replaces the natural-only information matrix
\(S^\top W_{yy,h}^{-1}S\) by
\(\Meas_S(U)^\top (W^S_h(U))^{-1}\Meas_S(U)\) in state coordinates, and
the resulting state covariance is mapped to the reported variables through
\(\Sigma^y_{0,h}=S\Sigma^S_{0,h}S^\top\) and
\(\Sigma^y_{U,h}=S\Sigma^S_{U,h}S^\top\). When \(S=I_n\), these reduce to
the ordinary multivariate covariance formulas.

\begin{corollary}[Covariance-risk reduction form of gain]
\label{cor:gain}
Let \(Q_h\succeq 0\) denote a horizon-specific weight matrix on the
reported variables. The unregularized population covariance-risk gain on the reported nodes is
\begin{equation}
\label{eq:ana-cov-risk-reduction}
\begin{aligned}
\mathcal G^{\mathrm{ana}}_{h,S}(U;Q_h)
&=
\operatorname{tr}\!\left(
Q_h(\Sigma^y_{0,h}-\Sigma^y_{U,h})
\right)\\
&=
\operatorname{tr}\!\left(
Q_hS
(\Sigma^S_{0,h}-\Sigma^S_{U,h})
S^\top
\right)\\
&=
\operatorname{tr}\!\left(
Q^b_h
(\Sigma^S_{0,h}-\Sigma^S_{U,h})
\right).
\end{aligned}
\end{equation}
Here \(Q^b_h=S^\top Q_hS\) is the state-coordinate weight induced by
\(Q_h\).

Hence the covariance-risk gain equals the reduction in weighted covariance
risk induced by the augmented projection system.
\end{corollary}

Through Equation~\eqref{eq:ana-cov-risk-reduction},
Corollary~\ref{cor:gain} turns auxiliary-direction design into a
covariance-risk reduction criterion on the reported target variables, or
equivalently on state coordinates using \(Q^b_h\).

\begin{proposition}[Population covariance-risk non-deterioration under correctly specified GLS augmentation]
\label{prop:psd}
Under Assumption~\ref{ass:wellposed}, define
\[
\begin{aligned}
T^S_h(U) &= R^S_h(U)-K^S_h(U)^\top W_{yy,h}^{-1}K^S_h(U), \\
M^S_h(U) &= U-S^\top W_{yy,h}^{-1}K^S_h(U).
\end{aligned}
\]
Here \(T^S_h(U)\) is the Schur complement of \(W_{yy,h}\) in the
joint residual covariance: it is the residual covariance of the auxiliary
forecast errors after the part linearly explained by the natural forecast
errors has been removed. The matrix \(M^S_h(U)\) is the corresponding
adjusted auxiliary direction matrix after the same natural-error
adjustment.
Then
\begin{equation}
\label{eq:info-matrix-decomp}
\begin{aligned}
\Meas_S(U)^\top (W^S_h(U))^{-1}\Meas_S(U)
&=
S^\top W_{yy,h}^{-1}S\\
&\quad
+ M^S_h(U)\big(T^S_h(U)\big)^{-1}M^S_h(U)^\top.
\end{aligned}
\end{equation}
Consequently,
\begin{equation}
\label{eq:population-nondeterioration}
\Sigma^S_{U,h}\preceq \Sigma^S_{0,h},
\qquad
\Sigma^y_{U,h}\preceq \Sigma^y_{0,h},
\qquad
\mathcal G^{\mathrm{ana}}_{h,S}(U;Q_h)\ge 0
\end{equation}
for every \(Q_h\succeq 0\). Equivalently, under correctly specified
covariances, augmentation cannot worsen the population covariance-risk
component of the unified GLS reconciliation system.
\end{proposition}

The information decomposition in Equation~\eqref{eq:info-matrix-decomp}
makes the population-level ideal in Proposition~\ref{prop:psd} explicit:
with correctly specified covariances and generalized least squares, adding
auxiliary measurements cannot worsen the covariance-risk component. This is
not a claim that augmentation cannot worsen MSE, realized quadratic risk, or
test loss; those quantities also depend on bias and finite-sample estimation.

\begin{lemma}[Quadratic-risk decomposition with bias]
\label{lem:bias-risk}
Let \(Q_h\succeq 0\) be conformable with the reported error vector. Define
the reported-error biases by
\[
\mu^y_{0,h}=\E[\varepsilon^y_{0,t,h}],
\qquad
\mu^y_{U,h}=\E[\varepsilon^y_{U,t,h}].
\]
Then the quadratic-risk improvement is
\begin{equation}
\label{eq:bias-risk-decomp}
\begin{aligned}
\mathcal G^{\mathrm{quad}}_{h,S}(U;Q_h) &=
\tr\!\big(Q_h(\Sigma^y_{0,h}-\Sigma^y_{U,h})\big)\\
&\quad
+ (\mu^y_{0,h})^\top Q_h \mu^y_{0,h}
- (\mu^y_{U,h})^\top Q_h \mu^y_{U,h}\\
&=
\mathcal G^{\mathrm{ana}}_{h,S}(U;Q_h)
+
(\mu^y_{0,h})^\top Q_h\mu^y_{0,h}
-
(\mu^y_{U,h})^\top Q_h\mu^y_{U,h}.
\end{aligned}
\end{equation}
\end{lemma}

\begin{remark}[Why the search criterion is direct gain]
The decomposition in Equation~\eqref{eq:bias-risk-decomp} shows that the
population quadratic-risk gain contains both a covariance-reduction
component and a bias-change component. Therefore, the covariance gain
\(\mathcal G^{\mathrm{ana}}_{h,S}\) is useful for explaining the variance-side
mechanism of augmentation, but it is not sufficient as the final
selection criterion. A direction with favorable covariance reduction
may still yield limited realized improvement if it increases bias.
Likewise, standalone auxiliary predictability cannot determine
reconciliation usefulness, because the direction matters only through
its net effect after augmented reconciliation. REGAIN therefore uses
the empirical loss-based search gain, which directly evaluates
the final target-weighted loss reduction.
\end{remark}

\subsection{Stability of estimated gain signals}
\label{sec:gain-signal-stability}

Having separated the direct quadratic-risk objective from its
covariance-side mechanism, we now examine the stability of that
covariance-side signal. In practice, the covariance matrices entering the
formula in Corollary~\ref{cor:gain} are estimated rather than known. The
question is whether the covariance-side gain remains close to its
population version when the population covariance family is replaced by its
estimated counterpart.

This stability result is not intended to replace the direct empirical
search criterion. Rather, it supports the use of gain-based information in
the search by showing that, under uniform covariance perturbations, the
estimated covariance-side gain remains close to its population version over
the normalized feasible set.

Let
\[
\mathcal U_k=\{U=D^{-1/2}V: V\in\St(n,k)\}.
\]
This is the normalized feasible image induced by the Stiefel
parameterization. Write the population information matrices as
\[
\begin{aligned}
\mathcal I^S_{0,h}
&=S^\top W_{yy,h}^{-1}S,\\
\mathcal I^S_h(U)
&=
\Meas_S(U)^\top
\big(W^S_h(U)\big)^{-1}
\Meas_S(U).
\end{aligned}
\]
Thus \(\Sigma^S_{0,h}=(\mathcal I^S_{0,h})^{-1}\) and
\(\Sigma^S_{U,h}=(\mathcal I^S_h(U))^{-1}\).
Given an estimated natural-measurement covariance \(\widehat W_{yy,h}\)
and an estimated augmented covariance family \(\widehat W^S_h(U)\), define,
whenever the covariance inverses exist, the estimated information matrices by
\[
\begin{aligned}
\widehat{\mathcal I}^S_{0,h}
&=S^\top \widehat W_{yy,h}^{-1}S,\\
\widehat{\mathcal I}^S_h(U)
&=
\Meas_S(U)^\top
\big(\widehat W^S_h(U)\big)^{-1}
\Meas_S(U).
\end{aligned}
\]
When these estimated information matrices are nonsingular, set
\[
\begin{aligned}
\widehat\Sigma^S_{0,h}
&=\big(\widehat{\mathcal I}^S_{0,h}\big)^{-1},\\
\widehat\Sigma^S_{U,h}
&=\big(\widehat{\mathcal I}^S_h(U)\big)^{-1},\\
\widehat\Sigma^y_{0,h}
&=S\widehat\Sigma^S_{0,h}S^\top,\\
\widehat\Sigma^y_{U,h}
&=S\widehat\Sigma^S_{U,h}S^\top .
\end{aligned}
\]
The corresponding estimated covariance-side gain is
\[
\widehat{\mathcal G}^{\mathrm{ana}}_{h,S}(U;Q_h)
=
\tr\!\left(
Q_h\big(\widehat\Sigma^y_{0,h}-\widehat\Sigma^y_{U,h}\big)
\right).
\]

\begin{assumption}[Uniform stability of the covariance family]
\label{ass:uniform-stability}
For each horizon $h$, there exist constants $\underline\lambda_h>0$,
$B_{\Meas,h}<\infty$, and $B_{Q,h}<\infty$ such that
\[
\begin{aligned}
\lambda_{\min}\!\big(W_{yy,h}\big)
&\ge \underline\lambda_h, \\
\inf_{U\in\mathcal U_k}\lambda_{\min}\!\big(W^S_h(U)\big)
&\ge \underline\lambda_h, \\
\sup_{U\in\mathcal U_k}\|\Meas_S(U)\|_{\mathrm{op}}
&\le B_{\Meas,h}, \\
\|Q_h\|_{\mathrm{op}}
&\le B_{Q,h}.
\end{aligned}
\]
Moreover, the estimated covariance matrices are symmetric and obey the
uniform perturbation bound
\[
\begin{aligned}
\|\widehat W_{yy,h}-W_{yy,h}\|_{\mathrm{op}}
&\le \delta_h,\\
\sup_{U\in\mathcal U_k}
\|\widehat W^S_h(U)-W^S_h(U)\|_{\mathrm{op}}
&\le \delta_h.
\end{aligned}
\]
\end{assumption}

Assumption~\ref{ass:uniform-stability} keeps the GLS covariance systems
uniformly well conditioned over the normalized search class and requires
the estimated natural-only and augmented covariance families to be
uniformly close to their population counterparts. These conditions allow
covariance perturbations to be transferred to perturbations of the
reconciled error covariance and then to perturbations of the
covariance-gain signal.

\begin{lemma}[Estimated covariance and gain perturbation]
\label{lem:perturb}
Under Assumptions~\ref{ass:wellposed}--\ref{ass:uniform-stability},
for each horizon \(h\) there are constants \(c_h>0\),
\(C_{\Sigma,h}<\infty\), and \(C_{G,h}<\infty\), depending only on fixed
problem dimensions and the conditioning and boundedness constants in the
assumptions, such that, whenever \(\delta_h\le c_h\), the estimated
covariance and information matrices appearing above are positive definite
uniformly over \(U\in\mathcal U_k\), and
\[
\begin{aligned}
&\|\widehat\Sigma^S_{0,h}-\Sigma^S_{0,h}\|_{\mathrm{op}}
+
\sup_{U\in\mathcal U_k}
\|\widehat\Sigma^S_{U,h}-\Sigma^S_{U,h}\|_{\mathrm{op}}\\
&\quad\le
C_{\Sigma,h}\delta_h .
\end{aligned}
\]
Moreover,
\[
\begin{aligned}
&\sup_{U\in\mathcal U_k}
\left|
\widehat{\mathcal G}^{\mathrm{ana}}_{h,S}(U;Q_h)
-
\mathcal G^{\mathrm{ana}}_{h,S}(U;Q_h)
\right|\\
&\quad\le
C_{G,h}\delta_h .
\end{aligned}
\]
\end{lemma}

\begin{corollary}[Horizon-aggregated covariance-gain stability]
\label{cor:uniform-gain}
Under the assumptions of Lemma~\ref{lem:perturb}, define
\[
\begin{aligned}
G_h(U)
&=\mathcal G^{\mathrm{ana}}_{h,S}(U;Q_h),\\
\widehat G_h(U)
&=\widehat{\mathcal G}^{\mathrm{ana}}_{h,S}(U;Q_h),
\end{aligned}
\]
and aggregate these horizon-level covariance gains as
\[
\begin{aligned}
G(U)
&=\sum_{h=1}^H \omega_h G_h(U),\\
\widehat G(U)
&=\sum_{h=1}^H \omega_h \widehat G_h(U).
\end{aligned}
\]
Then there exists a constant \(C<\infty\) such that
\[
\sup_{U\in\mathcal U_k}
\left|\widehat G(U)-G(U)\right|
\le
C\sum_{h=1}^H \omega_h \delta_h.
\]
\end{corollary}

\begin{remark}[Role of covariance-gain stability]
Lemma~\ref{lem:perturb} and Corollary~\ref{cor:uniform-gain} show that
the covariance-side gain signal is not an arbitrary proxy. When the
estimated natural-only and augmented covariance families are uniformly close
to their population counterparts,
the estimated covariance gain remains uniformly close to its population
counterpart over the whole normalized feasible set. Thus, the
variance-reduction component identified in
Section~\ref{sec:risk-interpretation-search-gain} can be stably
approximated through estimated covariance matrices.
\end{remark}

This stability result is deliberately limited to the covariance-side
signal. Together with Lemma~\ref{lem:bias-risk}, it explains why
covariance gain is useful as a stable mechanism but still not the
selection objective: REGAIN selects directions by empirical
\(\loss_Q\)-based direct gain as in Equation~\eqref{eq:empirical-direct-gain} on the search and selection splits, so
that candidate directions are evaluated by their net effect on final
target-weighted loss.

\subsection{Single-direction interpretation}
\label{sec:single-direction-interpretation}
The preceding subsections treat auxiliary measurements as a block and
separate covariance-side gain from direct empirical gain. The
single-direction case has a narrower role: it exposes the covariance
anatomy of one candidate direction. Throughout this subsection, \(k=1\)
and \(U=u\), with \(u\) viewed as an admissible normalized direction in
\[
\mathcal U_1=\{D^{-1/2}v:v\in\St(n,1)\}.
\]
The algebra below also applies to any nonzero direction for which the
population GLS systems in Assumption~\ref{ass:wellposed} are well posed.

The single-direction formula is therefore interpretive rather than
prescriptive. It explains the variance-reduction part of one direction,
while the actual search and selection steps still use empirical direct gain
so that bias changes and finite-sample effects are included in the
acceptance decision.

For \(k=1\), the auxiliary covariance block \(R^S_h(U)\) is the scalar
\(r^S_h(u)\), and the cross-covariance block \(K^S_h(U)\) is the vector
\(k^S_h(u)\). The scalar Schur complement, equivalently the
one-dimensional version of \(T^S_h(U)\),
\[
\tau^S_h(u)=r^S_h(u)-k^S_h(u)^\top W_{yy,h}^{-1}k^S_h(u)
\]
is strictly positive under Assumption~\ref{ass:wellposed}. It is the
residual variance of the auxiliary forecast error after removing the part
linearly explained by the natural measurement residuals.

\begin{proposition}[Single-direction interpretation]
\label{prop:single}
Under Assumption~\ref{ass:wellposed}, consider the case \(k=1\) with an
admissible auxiliary direction \(u\in\mathcal U_1\). Let
\(Q_h\succeq0\) be a reported-node weight matrix. Define
\[
\begin{aligned}
r^S_h(u)
&=\mathrm{Var}\!\big(e^{c,S}_{t,h}(u)\big),\\
k^S_h(u)
&=\mathrm{Cov}\!\big(e^y_{t,h},e^{c,S}_{t,h}(u)\big),\\
\tau^S_h(u)
&=r^S_h(u)-k^S_h(u)^\top W_{yy,h}^{-1}k^S_h(u),\\
\tilde u^S_h
&=u-S^\top W_{yy,h}^{-1}k^S_h(u),\\
x^S_h(u)
&=\Sigma^S_{0,h}\tilde u^S_h .
\end{aligned}
\]
As in Corollary~\ref{cor:gain}, let \(Q^b_h=S^\top Q_hS\) be the
state-coordinate weight induced by \(Q_h\). Then \(\tau^S_h(u)>0\), the
denominator below is positive, and the horizon-\(h\) covariance-risk gain is
\begin{equation}
\label{eq:single-direction-gain}
\begin{aligned}
\mathcal G^{\mathrm{ana}}_{h,S}(u;Q_h)
&=
\frac{
\big(x^S_h(u)\big)^\top Q^b_h x^S_h(u)
}{
\tau^S_h(u)+(\tilde u^S_h)^\top x^S_h(u)
}.
\end{aligned}
\end{equation}
\end{proposition}

Equation~\eqref{eq:single-direction-gain} gives the cleanest specialized
variance-side lens for what makes an auxiliary measurement useful through
covariance reduction. It comes from viewing the augmented information
matrix as a rank-one update of the baseline information matrix
\((\Sigma^S_{0,h})^{-1}\) and simplifying with the Woodbury identity.

\begin{remark}[Mechanism and scope of the single-direction formula]
The vector \(\tilde u^S_h\) is the auxiliary direction after removing the
part already explained through the natural forecast-error block. The vector
\(x^S_h(u)=\Sigma^S_{0,h}\tilde u^S_h\) is the baseline covariance exposure
seen along that adjusted direction. Thus the numerator in
Equation~\eqref{eq:single-direction-gain} measures target-weighted exposure,
whereas the denominator combines effective residual noise
\(\tau^S_h(u)\) with the baseline uncertainty
\((\tilde u^S_h)^\top x^S_h(u)\). A small residual-noise term does not by
itself create covariance gain if \(x^S_h(u)\) has little \(Q^b_h\)-weighted
mass; conversely, a target-exposed direction can be weakened by a large
residual-noise penalty. This mechanism does not account for bias reduction:
by Lemma~\ref{lem:bias-risk}, realized quadratic gain also depends on the
bias-change term. The empirical direct-gain criterion is therefore still
needed to evaluate net target-risk improvement.
\end{remark}

\begin{remark}[Relation to principal-component augmentation]
Under a linear-error specialization, let \(\xi_{t,h}\in\R^n\) be a latent
auxiliary-error vector such that
\(e^{c,S}_{t,h}(u)=u^\top \xi_{t,h}\). Define
\(R^S_h=\mathrm{Cov}(\xi_{t,h})\) and
\(K^S_h=\mathrm{Cov}(e^y_{t,h},\xi_{t,h})\). Then
\(r^S_h(u)=u^\top R^S_hu\) and \(k^S_h(u)=K^S_hu\), so
Proposition~\ref{prop:single} becomes a generalized Rayleigh quotient
\[
\mathcal G^{\mathrm{ana}}_{h,S}(u;Q_h)
=
\frac{u^\top A^S_h u}{u^\top C^S_h u},
\]
where
\[
\begin{aligned}
B^S_h
&=I-S^\top W_{yy,h}^{-1}K^S_h,\\
A^S_h
&=(B^S_h)^\top \Sigma^S_{0,h}Q^b_h\Sigma^S_{0,h}B^S_h,\\
C^S_h
&=R^S_h-(K^S_h)^\top W_{yy,h}^{-1}K^S_h\\
&\quad +(B^S_h)^\top \Sigma^S_{0,h}B^S_h.
\end{aligned}
\]
If \(C^S_h\succ0\), maximizing the single-direction analytic gain over
scale-free directions reduces to a leading generalized eigenvector problem
for \((A^S_h,C^S_h)\). Under the normalized parameterization
\(u=D^{-1/2}v\), the equivalent pair is
\[
\bigl(D^{-1/2}A^S_hD^{-1/2},
D^{-1/2}C^S_hD^{-1/2}\bigr).
\]
This clarifies the contrast with FLAP-style principal-component augmentation
\citep{Yang2024FLAP}: principal components rank directions by variance in a
fixed covariance matrix, whereas the analytic gain also depends on the target
weighting \(Q_h\), the natural measurement matrix \(S\), and the
complementary error term \(K^S_h\). Thus the point is not that principal
components cannot help, but that their ranking is not, by itself, a
target-aware covariance-gain criterion; the empirical direct-gain score is
still needed to evaluate net risk after bias effects.
\end{remark}

\section{Method}

Figure~\ref{fig:method-overview} summarizes the end-to-end \textsc{REGAIN}
pipeline. The notation and objective are those of
Section~\ref{sec:problem-setup}; this section focuses on how that objective is
scored and optimized in finite samples.

The theory in Section~\ref{sec:direct-gain-characterization} is
horizon-indexed: the population GLS projections use \(W_{yy,h}\) and
\(W^S_h(U)\) separately at each forecast horizon \(h\). The implementation
deliberately uses pooled stabilized estimators, denoted \(\widehat W_{yy}\)
and \(\widehat W^S(U)\), constructed from fit-split residuals across all
horizons. This pooling reduces covariance-estimation variance and stabilizes
finite-sample GLS. Thus \(\widehat W^S(U)\) should be read as a pooled
finite-sample proxy for the horizon-specific covariance family
\(\{W^S_h(U)\}_{h=1}^H\), not as a change in the theoretical target.

\subsection{Direct-gain scoring and covariance-stabilized surrogate}
\label{sec:direct-gain-scoring}

The computational core of \textsc{REGAIN} is a reusable scoring primitive.
Given a normalized direction matrix \(V\), set
\[
U_V=D^{-1/2}V .
\]
The primitive forecasts the auxiliary series induced by \(U_V\), estimates
the covariance matrices whose inverses enter GLS on the fit split, computes the
natural-only and augmented reconciled forecasts, and returns the
target-weighted loss reduction on a chosen scoring split.

The split roles are kept separate. The fit split \(\Tfit\) estimates the
covariance matrices used by reconciliation. A generic scoring split
\(\mathcal A\) only determines where the final loss reduction is averaged.
In the stagewise algorithm, \(\mathcal A=\Tsearch\) proposes directions and
\(\mathcal A=\Tselect\) screens them; \(\Ttest\) is reserved for final
reporting.

Let \(\widehat W_{yy}\succ0\) denote the stabilized natural-measurement
residual covariance estimated once from
\(\{e^y_{t,h}:t\in\Tfit,\ h=1,\ldots,H\}\) by the same
positive-definite shrinkage convention.
For a candidate \(U\), define the pooled augmented residual covariance
\[
\begin{aligned}
\widehat\Omega^S(U)
&=
\widehat{\operatorname{Cov}}
\left\{
\begin{bmatrix}
e^y_{t,h}\\
e^{c,S}_{t,h}(U)
\end{bmatrix}
\;:\;
\substack{t\in\Tfit\\ h=1,\ldots,H}
\right\}.
\end{aligned}
\]
The augmented covariance matrix used in GLS is the stabilized estimator
\begin{equation}
\label{eq:stabilized-covariance-weight}
\begin{aligned}
\widehat W^S(U)
&=
\bigl(1-\widehat\lambda(U)\bigr)\widehat\Omega^S(U)
+
\widehat\lambda(U)
\mathcal T\!\left(\widehat\Omega^S(U)\right),
\end{aligned}
\end{equation}
where \(\mathcal T(\cdot)\) is a structured positive-definite shrinkage
target. The estimated shrinkage intensity is computed on \(\Tfit\) and
truncated to \(\widehat\lambda(U)\in[\lambda_{\mathrm{sh}},1]\), for a
fixed small \(\lambda_{\mathrm{sh}}>0\). Since
\(\widehat\Omega^S(U)\succeq0\) and
\(\mathcal T(\widehat\Omega^S(U))\succ0\), this guarantees
\(\widehat W^S(U)\succ0\). This follows the logic of linear shrinkage
toward a structured positive-definite target
\citep{Ledoit2004Covariance,Schafer2005Shrinkage} and is in the spirit of
the shrinkage covariance estimators used in MINT-style reconciliation
\citep{Wickramasuriya2019MinT}. Consistent with the method-level convention
above, this pooled estimator is then used as the GLS weight for all forecast
horizons in the finite-sample scoring primitive.

The natural-only forecast used as the baseline is computed as
\[
\begin{aligned}
\tilde b^{(0)}_{t,h}
&\in
\arg\min_{b\in\R^n}\quad
\bigl(\hat y_{t,h}-Sb\bigr)^\top
\widehat W_{yy}^{-1}
\bigl(\hat y_{t,h}-Sb\bigr)\\
&\quad+\eta_{\mathrm{gls}}\|b\|_2^2,\\
\tilde y^{(0)}_{t,h}
&=S\tilde b^{(0)}_{t,h}.
\end{aligned}
\]
For the candidate \(U_V\), the augmented forecast is computed as
\[
\begin{aligned}
\tilde b_{t,h}(U_V)
&\in
\arg\min_{b\in\R^n}\quad
\bigl(\hat z^S_{t,h}(U_V)-\Meas_S(U_V)b\bigr)^\top
\bigl(\widehat W^S(U_V)\bigr)^{-1}\\
&\quad{}\times
\bigl(\hat z^S_{t,h}(U_V)-\Meas_S(U_V)b\bigr)
+\eta_{\mathrm{gls}}\|b\|_2^2,\\
\tilde y_{t,h}(U_V)
&=S\tilde b_{t,h}(U_V).
\end{aligned}
\]
For any scoring split \(\mathcal A\), the implemented direct-gain score is
\begin{equation}
\label{eq:implemented-direct-gain}
\begin{aligned}
\Gain_{\mathcal A}(V)
&=
\frac{1}{|\mathcal A|}
\sum_{t\in\mathcal A}
\sum_{h=1}^{H}
\omega_h
\Big[
\loss_Q\big(\tilde y^{(0)}_{t,h},y_{t+h}\big)
\Big.\\
&\qquad\Big.
-
\loss_Q\big(\tilde y_{t,h}(U_V),y_{t+h}\big)
\Big].
\end{aligned}
\end{equation}

Because \(\widehat W_{yy}\) and \(\widehat W^S(U_V)\) are always estimated
on \(\Tfit\), the same candidate matrix \(V\) can be evaluated on different
scoring splits without changing the reconciliation operator. In the
stagewise procedure below, we use
\[
\Gain_{\mathrm{search}}(V)=\Gain_{\Tsearch}(V),
\qquad
\Gain_{\mathrm{select}}(V)=\Gain_{\Tselect}(V).
\]
The search score is used to propose candidate directions, whereas the
selection score is used only to decide whether the proposed marginal gain
survives on held-out data.

The score \(\Gain_{\mathcal A}(V)\) in
Equation~\eqref{eq:implemented-direct-gain} is the re-evaluated direct-gain
score: the pooled covariance matrix \(\widehat W^S(U_V)\) is estimated at
the same candidate \(V\). During local numerical search, however,
re-estimating this covariance matrix after every trial gradient step can be
both expensive and unstable. We therefore use a frozen-covariance local
surrogate. Given an anchor point \(\bar V\), set
\[
\bar U=U_{\bar V}=D^{-1/2}\bar V,
\qquad
\bar W = \widehat W^S(\bar U),
\]
where \(\bar W\) is estimated on \(\Tfit\). For a nearby trial point \(V\),
define the frozen-covariance reconciled forecast
\[
\begin{aligned}
\widetilde b_{t,h}(V\mid \bar V)
&\in
\arg\min_{b\in\R^n}\quad
\bigl(\hat z^S_{t,h}(U_V)-\Meas_S(U_V)b\bigr)^\top
\bar W^{-1}\\
&\quad{}\times
\bigl(\hat z^S_{t,h}(U_V)-\Meas_S(U_V)b\bigr)
+\eta_{\mathrm{gls}}\|b\|_2^2 .
\end{aligned}
\]
Let \(\widetilde y_{t,h}(V\mid \bar V)=S\widetilde b_{t,h}(V\mid \bar V)\).
The corresponding surrogate score on a split \(\mathcal A\) is
\begin{equation}
\label{eq:frozen-covariance-surrogate}
\begin{aligned}
\widetilde{\Gain}_{\mathcal A}(V\mid \bar V)
&=
\frac{1}{|\mathcal A|}
\sum_{t\in\mathcal A}
\sum_{h=1}^H
\omega_h
\left[
\loss_Q\!\left(\tilde y^{(0)}_{t,h},y_{t+h}\right)
\right.\\
&\qquad\left.
-
\loss_Q\!\left(\widetilde y_{t,h}(V\mid \bar V),y_{t+h}\right)
\right].
\end{aligned}
\end{equation}
Only the pooled covariance matrix is frozen at the anchor point
\(\bar V\). The auxiliary series, auxiliary forecasts, measurement
matrix, and GLS solution are still recomputed at the trial point
\(V\). After the inner local steps, the resulting candidate is scored
again by the full \(\Gain_{\mathcal A}\), with the pooled covariance matrix
re-estimated at that candidate. Hence the surrogate affects only the
local proposal step, not the score used for stagewise comparison or
held-out acceptance.

\subsection{Stagewise \textsc{REGAIN}}
\label{sec:stagewise-regain}

The stagewise procedure is the core \textsc{REGAIN} estimator. Rather
than optimizing all auxiliary directions jointly from the start, it builds
the normalized direction matrix one column at a time using the scoring
primitive in Section~\ref{sec:direct-gain-scoring}. At every call to that
primitive, the actual auxiliary matrix is \(U_V=D^{-1/2}V\). The stagewise
construction avoids a high-dimensional joint search over \(\St(n,k)\) and
lets the method choose the direction dimension through held-out
marginal-gain checks. The full procedure is summarized in
Algorithm~\ref{alg:stagewise}.

At stage \(j\), suppose the already accepted normalized directions form
\(V_{j-1}\in\St(n,j-1)\), with \(V_0\) interpreted as the empty matrix.
For a trial direction \(v\), write
\[
V_{j-1}^{+}(v)=[V_{j-1},v].
\]
The feasible set for the next normalized direction is
\begin{equation}
\label{eq:stagewise-feasible-set}
\begin{aligned}
\mathcal C_j
&=
\left\{
v\in\R^n:
v^\top v=1,\;
V_{j-1}^\top v=0
\right\}.
\end{aligned}
\end{equation}
Thus \(V_{j-1}^{+}(v)\in\St(n,j)\) exactly when
\(v\in\mathcal C_j\). The search-split marginal gain of adding \(v\) is
\begin{equation}
\label{eq:stagewise-search-marginal}
\begin{aligned}
\Delta^{\mathrm{search}}_j(v)
&=
\Gain_{\mathrm{search}}\!\left(V_{j-1}^{+}(v)\right)
-
\Gain_{\mathrm{search}}(V_{j-1}).
\end{aligned}
\end{equation}
The conceptual stage-\(j\) proposal is therefore
\[
v_j^\star
\in
\operatorname*{arg\,max}_{v\in\mathcal C_j}
\Delta^{\mathrm{search}}_j(v),
\]
with the understanding that the implemented optimizer returns an approximate
maximizer.

To handle the orthogonality constraint explicitly, choose
\(R_{j-1}\in\R^{n\times(n-j+1)}\) such that
\[
\begin{aligned}
R_{j-1}^\top R_{j-1}&=I_{n-j+1},\\
V_{j-1}^\top R_{j-1}&=0,\\
R_{j-1}R_{j-1}^\top&=I_n-V_{j-1}V_{j-1}^\top .
\end{aligned}
\]
For \(j=1\), this convention gives \(R_0=I_n\). Every feasible direction
can then be written as
\[
v=R_{j-1}q,
\qquad
q\in\R^{n-j+1},
\qquad
\|q\|_2=1 .
\]
Define
\[
V_j(q)=V_{j-1}^{+}(R_{j-1}q).
\]
The marginal-gain search over \(\mathcal C_j\) is equivalently the
sphere-constrained problem
\begin{equation}
\label{eq:stagewise-sphere-objective}
\begin{aligned}
q_j^\star
&\in
\operatorname*{arg\,max}_{\|q\|_2=1}
\phi_j(q),\\
\phi_j(q)
&=
\Gain_{\mathrm{search}}\!\left(V_j(q)\right)
-
\Gain_{\mathrm{search}}(V_{j-1}).
\end{aligned}
\end{equation}
This parameterization separates the geometry from the scoring: the optimizer
only enforces a unit-sphere constraint in \(q\), while
\(R_{j-1}q\) is automatically orthogonal to the previously accepted
directions.

The full objective \(\phi_j\) is generally nonconvex because each evaluation
forecasts the induced auxiliary series, re-estimates the pooled covariance
matrix on \(\Tfit\), and solves the augmented GLS problem. For numerical
stability and efficiency, each restart uses the frozen-covariance surrogate
from Equation~\eqref{eq:frozen-covariance-surrogate} inside an outer--inner
loop. At an anchor \(\bar q\), set
\[
\bar V=V_j(\bar q),
\qquad
\bar U=U_{\bar V}=D^{-1/2}\bar V,
\qquad
\bar W=\widehat W^S(\bar U),
\]
where \(\bar W\) is estimated on \(\Tfit\). Holding this covariance matrix
fixed, the local surrogate marginal gain is
\begin{equation}
\label{eq:stagewise-surrogate-marginal}
\begin{aligned}
\widetilde\Delta^{\mathrm{search}}_j(q\mid \bar q)
&=
\widetilde{\Gain}_{\mathrm{search}}
\!\left(V_j(q)\mid \bar V\right)
-
\Gain_{\mathrm{search}}(V_{j-1}).
\end{aligned}
\end{equation}
The second term is constant in \(q\), so maximizing
\(\widetilde\Delta^{\mathrm{search}}_j(q\mid \bar q)\) is equivalent to
maximizing the frozen-covariance score of the trial matrix. After the inner
sphere steps, the trial point is re-evaluated by the full
\(\Delta^{\mathrm{search}}_j\), with the covariance matrix re-estimated at
the updated candidate. Across random restarts, the proposal \(q_j^\star\)
is chosen by the largest re-evaluated search-split marginal gain, and
\[
v_j^\star=R_{j-1}q_j^\star .
\]
This direction is only a search-split proposal; it is appended to
\(V_{j-1}\) only if it passes the held-out selection check.

The selection-split marginal gain of the proposal is
\begin{equation}
\label{eq:stagewise-selection-marginal}
\begin{aligned}
\Delta^{\mathrm{select}}_j(v_j^\star)
&=
\Gain_{\mathrm{select}}
\!\left(V_{j-1}^{+}(v_j^\star)\right)
-
\Gain_{\mathrm{select}}(V_{j-1}).
\end{aligned}
\end{equation}
The candidate is retained only if
\begin{equation}
\label{eq:stagewise-acceptance-rule}
\Delta^{\mathrm{select}}_j(v_j^\star)>\tau_j .
\end{equation}
If the test fails, the stagewise procedure stops and returns
\(V_{j-1}\). The threshold \(\tau_j\) controls how conservative the
held-out screening is; larger values require more selection-split evidence
before a new direction is accepted.

\begin{algorithm}[!t]
\caption{Stagewise Auxiliary Direction Learning}
\label{alg:stagewise}
\begin{algorithmic}[1]
\Require Direct-gain scores \(\Gain_{\mathrm{search}}\) and
\(\Gain_{\mathrm{select}}\), frozen-covariance surrogate
\(\widetilde{\Gain}_{\mathrm{search}}\), covariance-estimation split
\(\Tfit\), maximum direction count \(k_{\max}\le n\),
thresholds \(\{\tau_j\}_{j=1}^{k_{\max}}\)
\State Initialize \(V\gets[\,]\)
\For{\(j=1,\ldots,k_{\max}\)}
    \State Define
    \[
    \mathcal C_j
    =
    \{v\in\R^n:\ v^\top v=1,\; V^\top v=0\}.
    \]
    \State Define, for \(v\in\mathcal C_j\),
    \[
    \Delta^{\mathrm{search}}_j(v)
    =
    \Gain_{\mathrm{search}}([V,v])
    -
    \Gain_{\mathrm{search}}(V).
    \]
    \State Approximately solve
    \(v_j^\star\in
    \operatorname*{arg\,max}_{v\in\mathcal C_j}
    \Delta^{\mathrm{search}}_j(v)\)
    by multiple-restart outer--inner local search.
    \Statex \quad Outer steps re-estimate the pooled covariance matrix on
    \(\Tfit\).
    \Statex \quad Inner steps optimize
    \(\widetilde{\Gain}_{\mathrm{search}}\) with this matrix fixed.
    \Statex \quad Restart selection uses the re-evaluated
    \(\Gain_{\mathrm{search}}\).
    \State Compute
    \[
    \Delta^{\mathrm{select}}_j(v_j^\star)
    =
    \Gain_{\mathrm{select}}([V,v_j^\star])
    -
    \Gain_{\mathrm{select}}(V).
    \]
    \If{\(\Delta^{\mathrm{select}}_j(v_j^\star)>\tau_j\)}
        \State Accept \(v_j^\star\) and set \(V\gets[V,v_j^\star]\)
    \Else
        \State \textbf{break}
    \EndIf
\EndFor
\State \Return \(V\), the matrix of accepted normalized directions
\end{algorithmic}
\end{algorithm}

\begin{proposition}[Validation screening guarantee]
\label{prop:validation}
Fix a stagewise step \(j\) and the already accepted matrix
\(V_{j-1}\). Let \(\mathcal C_j\) be the feasible set in
Equation~\eqref{eq:stagewise-feasible-set}. In this proposition only, use
hats to distinguish empirical split scores from population functionals. Define
\[
\begin{aligned}
\Delta^{\mathrm{pop}}_j(v)
&=
\Gain\!\left(V_{j-1}^{+}(v)\right)
-
\Gain(V_{j-1}),\\
\widehat\Delta^{\mathrm{sel}}_j(v)
&=
\widehat{\Gain}_{\mathrm{select}}
\!\left(V_{j-1}^{+}(v)\right)
-
\widehat{\Gain}_{\mathrm{select}}(V_{j-1}).
\end{aligned}
\]
Here \(\Gain\) denotes the population direct-gain functional, while
\(\widehat{\Gain}_{\mathrm{select}}\) denotes the selection-split empirical
score written as \(\Gain_{\mathrm{select}}\) in the algorithm. Suppose that,
for some \(\varepsilon_{j,n}>0\) and \(\alpha_j\in(0,1)\),
\[
\Pr\!\left(
\sup_{v\in\mathcal C_j}
\left|
\widehat\Delta^{\mathrm{sel}}_j(v)
-
\Delta^{\mathrm{pop}}_j(v)
\right|
\le \varepsilon_{j,n}
\right)
\ge 1-\alpha_j .
\]
Let \(v_j^\star\) be any candidate returned by the search step. If it is
accepted according to Equation~\eqref{eq:stagewise-acceptance-rule}, namely
if
\[
\widehat\Delta^{\mathrm{sel}}_j(v_j^\star)>\tau_j
\qquad
\text{and}
\qquad
\tau_j>\varepsilon_{j,n},
\]
then
\[
\Pr\!\left(
\Delta^{\mathrm{pop}}_j(v_j^\star)\le 0
\;\text{and}\;
v_j^\star\ \text{is accepted}
\right)
\le \alpha_j .
\]
Equivalently, on the event that the selection-split marginal gain is
uniformly estimated within \(\varepsilon_{j,n}\), every accepted direction
has positive population marginal gain.
\end{proposition}

Proposition~\ref{prop:validation} upgrades the search/select split from a
heuristic safeguard to a model-selection mechanism with an explicit
statistical interpretation. The result does not require the search optimizer
to find the global maximizer of \(\Delta^{\mathrm{search}}_j\); the uniform
selection-split bound covers whichever candidate the search step returns.
Once the selection split estimates marginal gain uniformly well, the
acceptance rule with a sufficiently positive threshold controls the
probability of falsely accepting a direction whose population contribution is
nonpositive.

\FloatBarrier
\subsection{Optional joint refinement: \textsc{REGAIN-JR}}
\label{sec:regain-jr}

Algorithm~\ref{alg:jr} describes the optional joint refinement variant,
denoted \textsc{REGAIN-JR}. Given the stagewise solution
\(\widehat V_{\mathrm{stage}}\in\St(n,\widehat k)\), where
\(\widehat k\) is the number of accepted directions, the refinement keeps
\(\widehat k\) fixed and uses \(\widehat V_{\mathrm{stage}}\) only as a
local initialization. If \(\widehat k=0\), there is no direction to refine.
Otherwise, \textsc{REGAIN-JR} locally improves the same search-split
direct-gain score used in Section~\ref{sec:stagewise-regain}:
\begin{equation}
\label{eq:jr-local-objective}
\begin{aligned}
\max_{V\in\St(n,\widehat k)}
\quad
&\Gain_{\mathrm{search}}(V).
\end{aligned}
\end{equation}
This is not a second model-selection step. It does not add directions,
change \(\widehat k\), or rerun the held-out screening rule. Its role is to
let the already accepted directions move jointly within the selected
\(\widehat k\)-dimensional normalized direction class.

The forecasting oracle remains frozen throughout the refinement. All updates
are taken with respect to the normalized direction matrix \(V\), and the
unwhitened auxiliary matrix used by the scoring primitive is always
\[
U_V=D^{-1/2}V .
\]
Let \(F_U(U)\) denote a differentiable score written in unwhitened
coordinates and \(F(V)=F_U(U_V)\) its normalized-coordinate version. The
Euclidean chain rule gives
\begin{equation}
\label{eq:jr-chain-rule}
\begin{aligned}
\nabla_V F(V)
&=
\bigl(D^{-1/2}\bigr)^\top
\nabla_U F_U(U_V).
\end{aligned}
\end{equation}
Because \(D\) is taken to be symmetric positive definite, and diagonal in
the experiments, \((D^{-1/2})^\top=D^{-1/2}\). During the inner JR steps
below, \(F\) is the
frozen-covariance surrogate; hence the gradient is not taken through a
fresh covariance-estimation step at every inner trial.

Because \(V\) is constrained to \(\St(n,\widehat k)\), Euclidean gradients
are projected onto the Stiefel tangent space
\[
\begin{aligned}
T_V\St(n,\widehat k)
&=
\left\{
\Delta\in\R^{n\times\widehat k}:
V^\top\Delta+\Delta^\top V=0
\right\}.
\end{aligned}
\]
For an ambient gradient \(G_E=\nabla_VF(V)\), the embedded-metric projection
is
\begin{equation}
\label{eq:stiefel-gradient-projection}
\begin{aligned}
\operatorname{grad}F(V)
&=
\Pi_V(G_E)\\
&=
G_E
-
V\,\sym(V^\top G_E),
\qquad
\sym(A)=\frac{A+A^\top}{2}.
\end{aligned}
\end{equation}
A tangent step is mapped back to the Stiefel manifold by a retraction. We use
the QR retraction
\begin{equation}
\label{eq:qr-retraction}
\Retr_V(\Delta)=\qf(V+\Delta),
\end{equation}
where \(\qf(A)\) denotes the \(Q\)-factor of a QR factorization of \(A\),
with column signs fixed consistently
\citep{Edelman1998Geometry,Absil2008Optimization,Absil2007TrustRegion}.

As in Section~\ref{sec:direct-gain-scoring}, repeatedly re-estimating the
pooled covariance matrix inside each local step can be expensive and noisy.
We therefore use an outer--inner schedule. At outer iteration \(s\), take the
current matrix \(V^{(s)}\) as the anchor, set
\[
U_s=U_{V^{(s)}}=D^{-1/2}V^{(s)},
\qquad
\widehat W_s=\widehat W^S(U_s),
\]
and define the inner objective by the frozen-covariance surrogate
\begin{equation}
\label{eq:jr-frozen-objective}
\begin{aligned}
F_s(Z)
&=
\widetilde{\Gain}_{\mathrm{search}}
\!\left(Z\mid V^{(s)}\right).
\end{aligned}
\end{equation}
Only \(\widehat W_s\) is frozen. For a trial matrix \(Z\), the auxiliary
matrix \(U_Z\), auxiliary forecasts, measurement matrix, and GLS solution
are still recomputed at \(Z\), exactly as in
Equation~\eqref{eq:frozen-covariance-surrogate}. After the inner steps
produce \(V^{\mathrm{trial}}\), the trial is evaluated by the full
re-evaluated score \(\Gain_{\mathrm{search}}\), with the covariance matrix
estimated again at \(V^{\mathrm{trial}}\). The trial is accepted only if this
full score improves over the current value. Thus the frozen surrogate guides
local proposals, while final JR acceptance remains tied to the same
re-evaluated direct-gain score as the rest of \textsc{REGAIN}.

\begin{algorithm}[!tbp]
\caption{Optional Joint Gain Refinement}
\label{alg:jr}
\begin{algorithmic}[1]
\Require Stagewise initialization
\(V_0=\widehat V_{\mathrm{stage}}\in\St(n,\widehat k)\)
\Require Whitening matrix \(D\), outer limit \(N_{\mathrm{out}}\), inner steps \(L\)
\If{\(\widehat k=0\)}
  \State \Return \(V_0\)
\EndIf
\State Set \(V\leftarrow V_0\) and
\[
g\leftarrow \Gain_{\mathrm{search}}(V).
\]
\For{\(s=1,\ldots,N_{\mathrm{out}}\)}
  \State Set \(V^{(s)}\leftarrow V\),
  \(U_s\leftarrow D^{-1/2}V^{(s)}\), and
  \(\widehat W_s\leftarrow\widehat W^S(U_s)\) on \(\Tfit\)
  \State Define
  \[
  F_s(Z)
  =
  \widetilde{\Gain}_{\mathrm{search}}
  \!\left(Z\mid V^{(s)}\right).
  \]
  \State Initialize \(Z\leftarrow V^{(s)}\)
  \For{\(r=1,\ldots,L\)}
    \State Compute \(G_E\leftarrow\nabla_Z F_s(Z)\)
    \State Project \(G_R\leftarrow G_E-Z\,\sym(Z^\top G_E)\)
    \State Choose a step size \(\eta_{s,r}>0\)
    \State Update \(Z\leftarrow\Retr_Z(\eta_{s,r}G_R)\)
  \EndFor
  \State Set \(V^{\mathrm{trial}}\leftarrow Z\) and recompute
  \[
  g^{\mathrm{trial}}
  \leftarrow
  \Gain_{\mathrm{search}}(V^{\mathrm{trial}}).
  \]
  \If{\(g^{\mathrm{trial}}>g\)}
    \State Set \(V\leftarrow V^{\mathrm{trial}}\) and
    \(g\leftarrow g^{\mathrm{trial}}\)
  \EndIf
\EndFor
\State \Return \(V\)
\end{algorithmic}
\end{algorithm}

\FloatBarrier

\section{Empirical Evaluation}

This section evaluates whether gain-selected auxiliary directions improve
final forecast accuracy when the forecasting oracle, covariance-estimation
split, and reconciliation rule are held fixed. We use two complementary
benchmarks. Beijing PM2.5 is an ordinary multivariate task with
\(S=I_n\), so any improvement must come from learned linear measurements of
the observed station series. Tourism is a structured hierarchy with a supplied
aggregation matrix \(S\), so it tests whether learned auxiliary measurements
can add value beyond standard no-auxiliary hierarchical reconciliation. The
evaluation compares gain-selected directions with no-auxiliary reconciliation,
fixed or random auxiliary baselines, and the optional joint-refinement variant.

\subsection{Benchmarks and Forecasting Protocol}
\label{sec:data-protocol}

We consider two complementary benchmark settings.
\begin{itemize}
    \item \textbf{Beijing PM2.5.} This is the ordinary multivariate benchmark. We form a daily data set by aggregating hourly PM2.5 measurements from the Beijing Multi-Site Air-Quality data across monitoring stations \citep{Zhang2017CautionaryBeijing}. Forecast accuracy is evaluated directly on the observed station-level PM2.5 series.
    \item \textbf{Tourism.} This is the natural-hierarchy benchmark, which contains monthly Australian tourism observations from January 1998 to December 2017. We use its given aggregation structure to test whether learned auxiliary series outside the supplied hierarchy can provide useful additional information and further improve accuracy on the original hierarchy levels \citep{Athanasopoulos2011TourismCompetition}.
\end{itemize}

We divide the rolling forecast origins into four disjoint segments. The fit
segment \(\Tfit\) estimates the pooled joint error covariance. The search
segment \(\Tsearch\) optimizes the gain objective for candidate directions.
The selection segment \(\Tselect\) accepts or rejects proposed directions and
determines when the stagewise phase stops. The test segment \(\Ttest\) is
used only for final reporting. Thus covariance fitting, direction search,
held-out screening, and test evaluation use separate origins.
The forecasting horizons and rolling-origin partitions are summarized in Table~\ref{tab:rolling_protocol}.

\vspace{4pt}
\noindent
\begin{minipage}{\columnwidth}
\captionof{table}{Rolling-origin forecasting protocol.}
\label{tab:rolling_protocol}
\centering
\footnotesize
\setlength{\tabcolsep}{3pt}
\begin{tabular}{@{}lcccc@{}}
\toprule
Data & Context $L$ & Seasonality & $H$ & Split \\
\midrule
Beijing PM2.5 & 30 & 7 & 7 & 761/218/139/286 \\
Tourism & 24 & 12 & 8 & 77/52/28/28 \\
\bottomrule
\end{tabular}

\vspace{2pt}
\raggedright
\footnotesize
\emph{Note}: The split column gives the number of rolling origins in
$|\Tfit|/|\Tsearch|/|\Tselect|/|\Ttest|$.
\end{minipage}
\vspace{4pt}

The forecasting model is treated as an external component shared across all
compared methods. We consider independent forecasting and, where available,
joint forecasting by a shared multivariate oracle. All comparisons are made
within the same dataset, oracle family, and forecast mode, so differences are
attributable to the auxiliary-direction design and reconciliation step rather
than to changes in the forecasting engine. The shared forecasters are
instantiated with off-the-shelf time-series foundation-model families such as
Chronos and TimesFM \citep{Ansari2024Chronos,Das2024TimesFM}; implementation
details and the available forecast modes are given in Appendix~\ref{app:frozen_oracles}.

\subsection{Experimental Configuration}
\label{sec:experimental-configuration}

Under the protocol in Section~\ref{sec:data-protocol}, we use the following
common configuration for all auxiliary-direction methods.

Let \(m\) denote the number of reported variables on which the final loss is
evaluated: the 12 station series for Beijing PM2.5 and all reported hierarchy
nodes for Tourism. Let \(\varsigma_i^2\) be the empirical variance of the
\(i\)-th evaluation variable, estimated on \(\Tfit\). We set
\[
Q=\diag\!\left(
\frac{1}{\varsigma_1^2+10^{-8}},\dots,
\frac{1}{\varsigma_m^2+10^{-8}}
\right)
\]
and use uniform horizon weights \(\omega_h=1/H\). Thus the gain score is a
scale-normalized loss reduction on the same reported variables used for final
evaluation.

All learned and random auxiliary directions are represented in whitened
coordinates. Let \(x_t\) be the state series on which auxiliary directions are
parameterized: \(x_t=y_t=b_t\) for Beijing PM2.5 and \(x_t=b_t\), the
bottom-level tourism series, for Tourism. Let \(\widehat\Sigma_x\) be the
covariance of \(x_t\) estimated on \(\Tfit\), and let
\(s_i^2=\widehat\Sigma_{x,ii}\). We use
\[
D=\operatorname{diag}(d_1,\ldots,d_n),
\qquad
d_i=\max\{s_i^2,\epsilon_D\},
\]
where \(s_i^2\) is the empirical variance of the \(i\)-th series and
\[
\epsilon_D
=
\max\left\{
10^{-8},\,
10^{-4}\operatorname{median}\{s_j^2:s_j^2>0\}
\right\}.
\]
Candidate directions are optimized in normalized coordinates and mapped back by
\[
U=D^{-1/2}V,
\qquad
V^\top V=I .
\]

For covariance estimation, we fit a single Ledoit--Wolf shrinkage
covariance estimator \citep{Ledoit2004Covariance} using residuals pooled
over \(\Tfit\) and all forecast horizons. This pooled-horizon estimate
avoids separate horizon-wise covariance fitting and is used uniformly
across horizons in the corresponding reconciliation solve.
 
For each natural-only or augmented reconciliation solve, we add a small
scale-adaptive ridge term to the GLS normal matrix. Let
\[
A_{\mathrm{gls}}=\Meas^\top W^{-1}\Meas, \qquad
\eta_{\mathrm{gls}}
=
10^{-6}\cdot
\frac{\tr(A_{\mathrm{gls}})}{n}.
\]
Here \(\Meas\) denotes the relevant natural-only or augmented measurement
matrix in the ordinary multivariate or structured setting, and \(W\) denotes
the corresponding stabilized reconciliation weight matrix used in that solve. Since
\(\tr(A_{\mathrm{gls}})/n\) is the average curvature scale of the GLS normal matrix,
the ridge term is adaptive to the scale of the current solve while remaining a
small numerical safeguard. This choice follows the standard ridge/Tikhonov regularization principle
\citep{hoerl1970ridge,golub1999tikhonov} and is analogous in spirit to
curvature-scaled damping in Levenberg--Marquardt-type methods
\citep{marquardt1963algorithm}.

The stagewise acceptance budget \(k_{\max}\) is varied over
\(\{1,2,4,6,8\}\), while the realized number of accepted directions,
denoted by \(\widehat k\), is determined by the selection-split
marginal-gain rule and satisfies \(0\le \widehat k\le k_{\max}\).
Each new direction is searched from 8 random initializations on the
feasible sphere. For each restart, we optimize the fixed-covariance
surrogate by Riemannian gradient ascent, where the Euclidean gradient
obtained by automatic differentiation is projected onto the tangent space
of the feasible sphere and then retracted back by normalization. This local
search is run for 3 outer loops with 5 inner updates per loop, using a step
size of 0.05.

For the stagewise acceptance rule, we set \(\tau_j=0\) for all stages \(j\).
Thus a proposed direction is retained whenever its selection-split marginal
direct gain is positive. The finite-sample screening guarantee in
Proposition~\ref{prop:validation} applies to the more conservative regime
\(\tau_j>\varepsilon_{j,n}\); the zero-threshold choice used in the experiments
is an empirical acceptance rule rather than an invocation of that guarantee.

When the optional \textsc{REGAIN-JR} phase is used, we apply a local
Riemannian refinement to the stagewise solution, with the pooled covariance
matrix re-estimated between outer updates. The refinement uses at most
10 outer iterations and 5 inner fixed-covariance steps per outer iteration.
A trial point is accepted only when the recomputed search gain improves over
the current value.

These choices are implementation safeguards rather than conceptual ingredients
of the method, and they are kept fixed across datasets unless explicitly stated
otherwise.

\subsection{Baselines and Comparisons}

We use \textsc{REGAIN} for the stagewise procedure and
\textsc{REGAIN-JR} for the variant with the optional refinement phase. The
no-auxiliary comparator is dataset dependent. For Beijing PM2.5, where
\(S=I_n\), the no-auxiliary reconciled forecast coincides with the direct
station-level base forecast. For Tourism, the no-auxiliary comparator is
\textsc{MINT} reconciliation under the supplied hierarchy; the unreconciled
\textsc{BASE} forecast is reported only as a reference for the value of
standard hierarchical reconciliation.

The main comparison groups are as follows.
\begin{enumerate}
    \item Direct base forecasts without auxiliary directions (\textsc{BASE}), reported for the original evaluation variables.
    \item For hierarchical benchmarks, no-auxiliary MINT reconciliation with the supplied aggregation matrix (\textsc{MINT}).
    \item Random auxiliary directions followed by augmented projection or reconciliation (\textsc{RAND}).
    \item Principal-component augmentation (\textsc{FLAP}). This baseline forms fixed auxiliary measurements from principal-component directions and augments the original system for joint projection or reconciliation
\citep{Yang2024FLAP}.
    \item Predictability-based auxiliary selection (\textsc{PRED}), where auxiliary aggregates are ranked by the forecasting quality of the auxiliary series themselves.
    \item Direct candidate-wise gain optimization with stagewise direction discovery (\textsc{REGAIN}).
    \item \textsc{REGAIN} followed by optional joint refinement (\textsc{REGAIN-JR}).
\end{enumerate}
After a direction set is chosen, all auxiliary-direction methods are evaluated
through the same covariance estimator and augmented GLS solve. The baseline
differences are therefore in how the auxiliary directions are selected, not in
the final reconciliation operator.

\subsection{Evaluation Metrics and Main Results}

The primary metric is target-weighted test gain on the reported evaluation
nodes. Let \(\tilde y^{(0)}_{t,h}\) denote the no-auxiliary comparator in the
corresponding dataset and oracle setting: direct station-level base forecasts
for Beijing PM2.5 and no-auxiliary \textsc{MINT} for Tourism. For a method
\(\mathsf M\), let \(\tilde y^{\mathsf M}_{t,h}\) be its final forecast on the
same reported nodes. We report
\[
\begin{aligned}
\Gain_{\mathrm{test}}(\mathsf M)
&=
\frac{1}{|\Ttest|}
\sum_{t \in \Ttest}
\sum_{h=1}^H
\omega_h
\Big[
\loss_Q\big(\tilde y^{(0)}_{t,h},y_{t+h}\big)\\
&\qquad
-
\loss_Q\big(\tilde y^{\mathsf M}_{t,h},y_{t+h}\big)
\Big].
\end{aligned}
\]
Positive values mean that the method improves on the dataset-specific
no-auxiliary comparator in target-weighted loss. For Tourism, the unreconciled
\textsc{BASE} row can therefore have negative gain because it is compared
against \textsc{MINT}. Main results are reported separately under independent
and joint base-forecast settings. More granular forecast-horizon and
node-level distributional breakdowns are used as diagnostics.

For Tourism, we report bottom-level RMSE, upper-level aggregate RMSE, all-node
RMSE, and all-node gain. For Beijing PM2.5, which has no supplied hierarchy,
we do not use level-wise metrics. Instead, we report MAE, RMSE, and gain on
the observed station series themselves; any grouping-based summaries are
diagnostic rather than part of the benchmark definition.

\begin{table*}[t]
	\centering
	\caption{Beijing PM2.5 results under the available base-forecast settings. We report MAE, RMSE, and test gain on the observed station-level PM2.5 series.}
	\label{tab:beijing-results}
	\scriptsize
	\setlength{\tabcolsep}{2.4pt}
	\renewcommand{\arraystretch}{0.96}
	\begin{tabular*}{\textwidth}{@{\extracolsep{\fill}}lccccccccc}
		\toprule
		\multirow{2}{*}{Method} & \multicolumn{3}{c}{Chronos2 indep.} & \multicolumn{3}{c}{Chronos2 joint} & \multicolumn{3}{c}{TimesFM indep.} \\
		\cmidrule(lr){2-4}\cmidrule(lr){5-7}\cmidrule(lr){8-10}
		& MAE $\downarrow$ & RMSE $\downarrow$ & Gain $\uparrow$
		& MAE $\downarrow$ & RMSE $\downarrow$ & Gain $\uparrow$
		& MAE $\downarrow$ & RMSE $\downarrow$ & Gain $\uparrow$ \\
		\midrule
		\textsc{BASE} & 51.98$\pm$0.00 & 74.69$\pm$0.00 & -0.00$\pm$0.00 & 51.85$\pm$0.00 & 74.50$\pm$0.00 & -0.00$\pm$0.00 & 51.60$\pm$0.00 & 75.30$\pm$0.00 & 0.00$\pm$0.00 \\
        \textsc{RAND} & 51.87$\pm$0.07 & 74.57$\pm$0.16 & 0.05$\pm$0.07 & 51.52$\pm$0.18 & 74.00$\pm$0.36 & 0.20$\pm$0.15 & 51.90$\pm$1.10 & 75.89$\pm$1.50 & -0.25$\pm$0.64 \\
        \textsc{FLAP} & 52.07$\pm$0.00 & 74.60$\pm$0.00 & 0.03$\pm$0.00 & 51.88$\pm$0.00 & 74.37$\pm$0.00 &  0.06$\pm$0.00 &  51.63$\pm$0.00 & 75.17$\pm$0.00 & 0.05$\pm$0.00 \\
        \textsc{PRED} & 51.89$\pm$0.05 & 74.37$\pm$0.09 & 0.13$\pm$0.04 & 51.77$\pm$0.04 & 74.25$\pm$0.07 & 0.10$\pm$0.03 &51.73$\pm$0.18 & 75.55$\pm$0.37 & -0.10$\pm$0.15 \\
        \textsc{REGAIN} & 51.77$\pm$0.11 & 74.30$\pm$0.20 & 0.16$\pm$0.08 & \textbf{51.49$\pm$0.13} & \textbf{73.92$\pm$0.19} & \textbf{0.23$\pm$0.08} & 51.48$\pm$0.10 & 75.07$\pm$0.24 & 0.09$\pm$0.10 \\
        \textsc{REGAIN-JR} & \textbf{51.75$\pm$0.13} & \textbf{74.25$\pm$0.24} & \textbf{0.18$\pm$0.10} & 51.65$\pm$0.11 & 74.09$\pm$0.14 & 0.16$\pm$0.06 & \textbf{51.46$\pm$0.12} & \textbf{75.06$\pm$0.25} & \textbf{0.10$\pm$0.10} \\

		\bottomrule
	\end{tabular*}
	\renewcommand{\arraystretch}{1.0}
	\par\vspace{2pt}
	\parbox{\linewidth}{\footnotesize Note: MAE, RMSE, and Gain are reported as mean$\pm$standard deviation over six seeds on the observed station-level PM2.5 series. Gain is measured relative to the no-auxiliary baseline under the same oracle and forecast mode; hence the \textsc{BASE} row has zero gain by definition. Negative gain indicates worse target-weighted loss than the corresponding no-auxiliary baseline. \textbf{Bold} marks the best entry within each setting.
}
\end{table*}

\begin{table*}[t]
	\centering
	\caption{Tourism results under the available base-forecast settings. We report bottom-level, upper-level, and all-node RMSE together with test gain; Figure~\ref{fig:tourism-k-sweep} shows the full direction-budget sweep curves.}
	\label{tab:main-results}
	\scriptsize
	\setlength{\tabcolsep}{2.4pt}
	\renewcommand{\arraystretch}{0.96}
	\resizebox{\textwidth}{!}{%
	\begin{tabular}{lcccccccccccc}
		\toprule
		\multirow{2}{*}{Method} & \multicolumn{4}{c}{Chronos2 indep.} & \multicolumn{4}{c}{Chronos2 joint} & \multicolumn{4}{c}{TimesFM indep.} \\
		\cmidrule(lr){2-5}\cmidrule(lr){6-9}\cmidrule(lr){10-13}
		& Bottom $\downarrow$ & Upper $\downarrow$ & ALL $\downarrow$ & Gain $\uparrow$
		& Bottom $\downarrow$ & Upper $\downarrow$ & ALL $\downarrow$ & Gain $\uparrow$
		& Bottom $\downarrow$ & Upper $\downarrow$ & ALL $\downarrow$ & Gain $\uparrow$ \\
		\midrule
		\textsc{BASE} & 209.8 & 1296.8 & 748.6$\pm$0.0 & -10.06$\pm$0.00 & 195.7 & 1260.9 & 726.3$\pm$0.0 & -3.16$\pm$0.00 & 215.6 & 1360.8 & 784.7$\pm$0.0 & -8.66$\pm$0.00 \\
        \textsc{MINT} & 202.2 & 1277.9 & 736.8$\pm$0.0 & 0.00$\pm$0.00 & 192.8 & 1244.6 & 716.8$\pm$0.0 & 0.00$\pm$0.00 & 209.4 & 1353.3 & 779.4$\pm$0.0 & 0.00$\pm$0.00 \\
        \textsc{RAND} & 202.1 & 1277.6 & 736.6$\pm$0.8 & 0.02$\pm$0.90 & 192.5 & 1243.1 & 715.9$\pm$1.7 & 0.07$\pm$0.99 & 209.4 & 1352.5 & 779.0$\pm$0.9 & -0.48$\pm$0.85 \\
        \textsc{FLAP} & 201.2 & 1261.6 & 730.7$\pm$0.0 & 0.46$\pm$0.00 & 192.1 & 1237.9 & 713.1$\pm$0.0 & 0.26$\pm$0.41 & 209.5 & 1352.7 & 779.8$\pm$0.0 & -0.59$\pm$0.00 \\
        \textsc{PRED} & 202.1 & 1277.5 & 736.8$\pm$3.2 & -0.21$\pm$2.02 & 192.7 & 1244.1 & 716.6$\pm$4.7 & -0.57$\pm$3.86 & 209.4 & 1352.2 & 778.8$\pm$0.9 & 0.19$\pm$0.54 \\
        \textsc{REGAIN} & 201.1 & 1257.4 & 724.9$\pm$3.7 & 2.29$\pm$1.15& 180.4 & 1170.7 & 674.3$\pm$11.1 & 9.61$\pm$4.55 & 209.1 & 1351.2 & 777.6$\pm$2.7 & 0.23$\pm$1.12 \\
	        \textsc{REGAIN-JR} & \textbf{201.0} & \textbf{1257.2} & \textbf{724.3$\pm$3.7} & \textbf{2.31$\pm$1.16} & \textbf{180.4} & \textbf{1170.4} & \textbf{673.9$\pm$11.4} & \textbf{9.68$\pm$4.60} & \textbf{208.9} & \textbf{1350.1} & \textbf{776.9$\pm$2.6} & \textbf{0.25$\pm$1.13} \\
		\bottomrule
	\end{tabular}
	}
	\renewcommand{\arraystretch}{1.0}
	\par\vspace{2pt}
	\parbox{\linewidth}{\footnotesize Note: Bottom and Upper report seed means only, while ALL and Gain are reported as mean$\pm$standard deviation over six seeds. Gain is measured relative to the no-auxiliary MINT reconciliation under the same oracle and forecast mode; hence the MINT row has zero gain by definition. The BASE row is unreconciled reference. Negative Base gain indicates worse target-weighted loss than the corresponding MINT baseline. \textbf{Bold} marks the best entry within each setting.}
\end{table*}

\begin{figure*}[t!]
\centering
\includegraphics[width=1\linewidth]{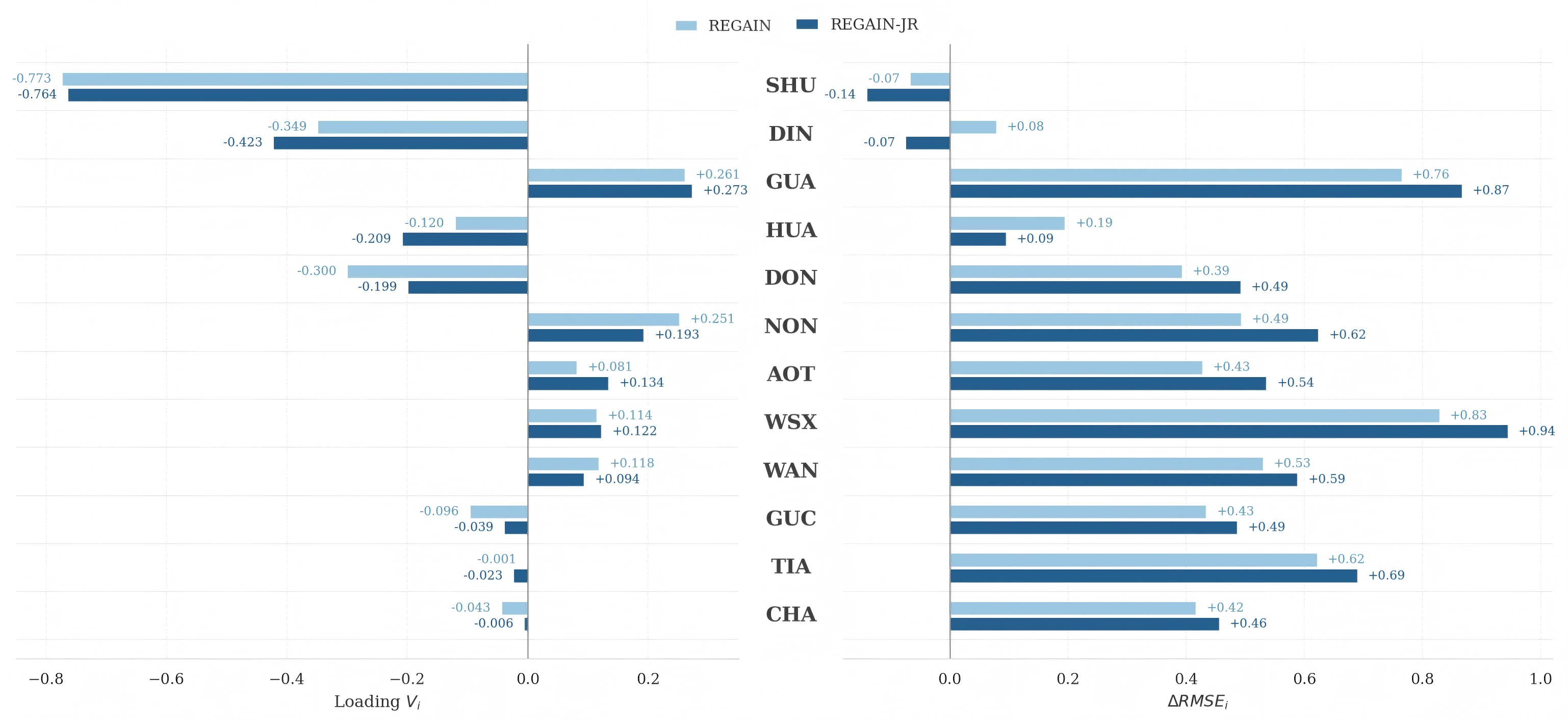}
\caption{Station-level interpretation on Beijing PM2.5. The left panel shows station-wise loadings of the accepted auxiliary direction, and the right panel shows station-wise RMSE improvement relative to the corresponding no-auxiliary base forecast. Positive \(\Delta \mathrm{RMSE}_i\) means that the auxiliary method reduces RMSE at station \(i\). REGAIN-JR largely preserves the stagewise pattern and yields only limited additional change.}
\label{fig:pm25-direction-interpretation}
\end{figure*}

\begin{figure*}[t]
	\centering
	\includegraphics[width=0.96\linewidth]{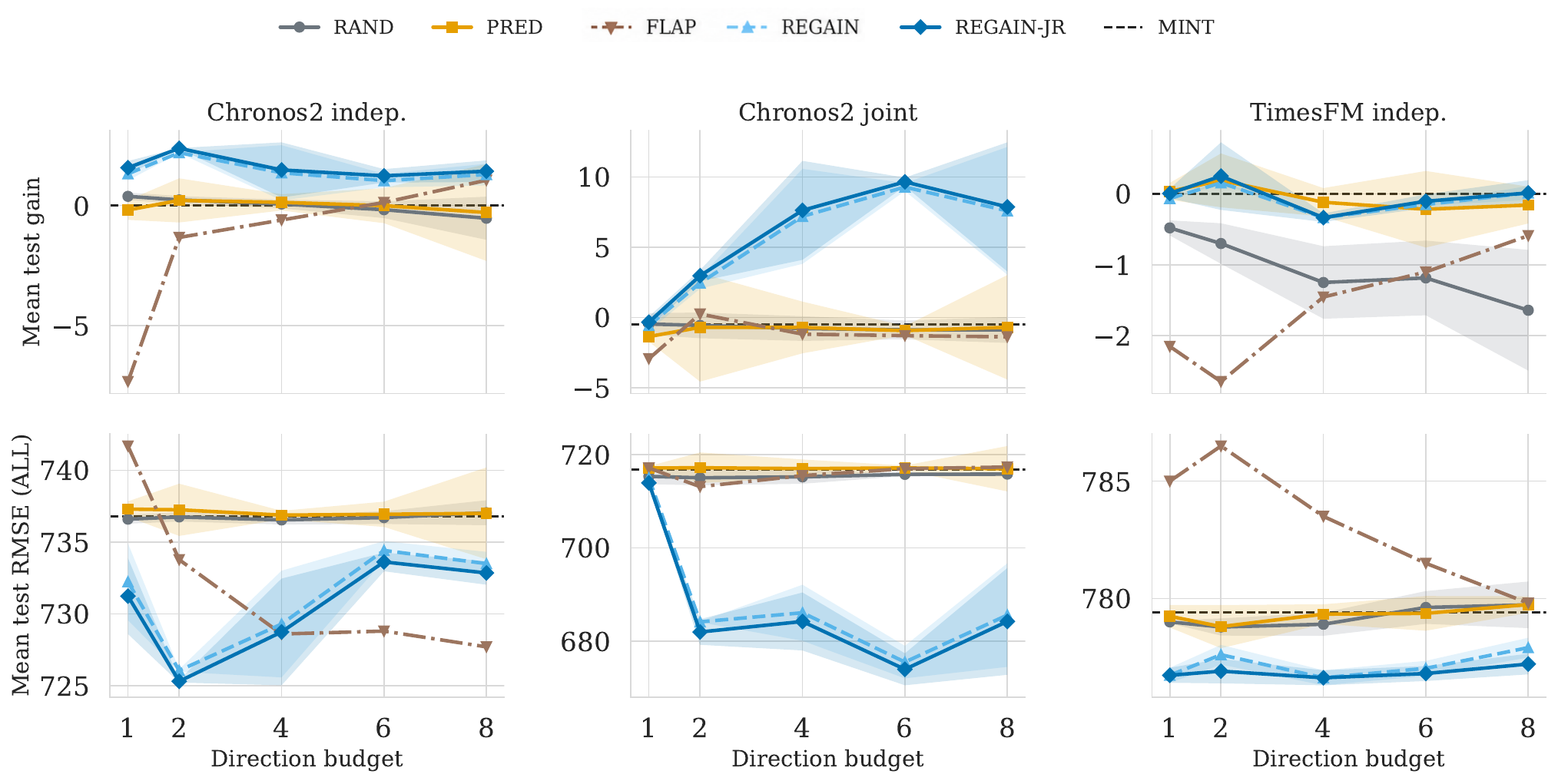}
    \caption{Tourism direction-budget sweep across predictor and forecast-mode settings. Curves show seed means with one-standard-deviation bands; the top row reports test gain and the bottom row reports overall test RMSE as the displayed direction budget varies. For RAND, FLAP, and PRED, this budget is the fixed number of auxiliary directions \(k\). For REGAIN and REGAIN-JR, it is the stagewise acceptance cap \(k_{\max}\). The no-auxiliary \textsc{MINT} baseline is shown as a horizontal reference.}
	\label{fig:tourism-k-sweep}
\end{figure*}

Tables~\ref{tab:beijing-results} and~\ref{tab:main-results} show that
gain-selected auxiliary directions can improve final forecast accuracy under
shared frozen-oracle base forecasts, but the size of the improvement depends
on the dataset and forecast mode. On Beijing PM2.5, \textsc{REGAIN} and
\textsc{REGAIN-JR} usually reduce station-level RMSE and yield positive test
gain relative to the no-auxiliary base forecast. This supports the ordinary
multivariate use case: useful auxiliary measurements can be learned even when
there is no supplied hierarchy. On Tourism, the strongest improvement occurs
under Chronos2 joint forecasting, where learned auxiliary directions
substantially reduce all-node RMSE relative to no-auxiliary \textsc{MINT}.
Chronos2 independent forecasting also gives positive mean gains, whereas the
TimesFM independent setting yields only small mean gains relative to
seed-to-seed variability. The bottom- and upper-level RMSEs decrease together
for the selected \textsc{REGAIN} variants, suggesting that the learned
measurements add information to the hierarchy rather than merely shifting
error between levels.

The baseline comparisons point to the same broad conclusion across the two
tables: selecting directions by downstream gain is more reliable in these
experiments than selecting them by standalone auxiliary predictability or by a
fixed variance-explained component rule. \textsc{PRED} can choose auxiliary
series that are easy to forecast but weak for reconciliation, and
\textsc{FLAP} can be useful in some settings but is less stable across
forecast modes. \textsc{REGAIN-JR} is not uniformly better than
\textsc{REGAIN} on the test split, which is expected because its trial
acceptance is based on recomputed search gain rather than test gain; its role
is local post-processing, not a new selection mechanism.

Figure~\ref{fig:pm25-direction-interpretation} shows that the learned auxiliary direction on Beijing PM2.5
has a nontrivial contrast structure across stations rather than behaving
like a uniform global average. The resulting RMSE improvements are highly
heterogeneous: most gains are concentrated on a subset of stations,
while a few stations improve only marginally or slightly deteriorate.
REGAIN-JR preserves the same overall pattern and provides only limited
additional gains beyond the stagewise core.

As a compact summary of the Tourism budget sweep,
Figure~\ref{fig:tourism-k-sweep} shows how the displayed direction budget
translates into test gain and overall RMSE across the available oracle and
forecast-mode settings. The sweep reveals three recurring patterns. First,
the direct-gain methods generally outperform random directions and
predictability-only selection at comparable budgets in the reported settings.
Under this controlled budget comparison, \textsc{FLAP} is less competitive;
part of that gap may reflect that all methods are aligned to the same
auxiliary-direction counts, which is useful for comparison but not necessarily
the most favorable operating regime for a fixed principal-component baseline.
Second, the useful operating region is moderate rather than monotone:
Chronos2 independent forecasting improves most clearly around the middle of
the grid, Chronos2 joint forecasting reaches high gains with only a small
number of accepted directions, and TimesFM independent forecasting requires a
larger cap before the gain becomes clearly positive. Third, \textsc{REGAIN}
and \textsc{REGAIN-JR} are close across the sweep, indicating that the
stagewise core captures most of the improvement. The optional refinement adds
only marginal extra benefit, mainly when the stagewise solution has already
found a transferable direction set and the remaining adjustment can be made
locally, as in the Chronos2 joint and TimesFM independent settings.

\subsection{Diagnostic Analyses}

We use diagnostics to explain why the learned directions help: marginal gain
per accepted direction, the relationship between gain and standalone auxiliary
predictability, the roles of the auxiliary residual covariance
\(R^S_h(U)\) and cross-covariance \(K^S_h(U)\), the structure of the learned
loadings, and how gains are distributed across horizons and target variables.
To keep the diagnostic block readable, the detailed figures below are shown
for Tourism under Chronos2 joint forecasting, the setting with the largest
\textsc{REGAIN} test gain among the reported base-forecast regimes. These
figures are explanatory diagnostics rather than additional model-selection
criteria.

Together, the diagnostic figures clarify the empirical mechanism behind the
aggregate results: how the stagewise procedure stops, why predictability alone
is insufficient, what drives realized gain, what the learned directions look
like, and how gains are distributed across horizons and target variables.

\begin{figure}[!t]
\centering
\includegraphics[width=\columnwidth]{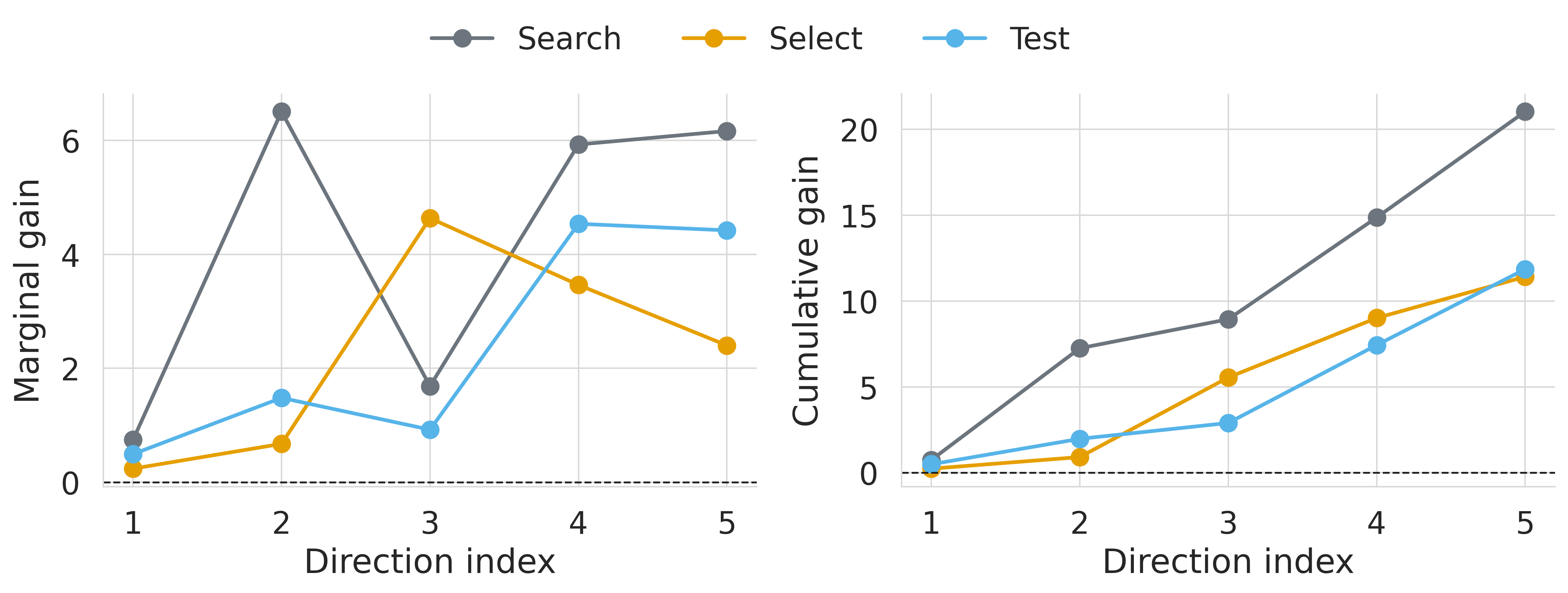}
\caption{Tourism marginal and cumulative gain paths. The figure reports the incremental and cumulative gain of accepted directions on the search, selection, and test segments.}
\label{fig:marginal-gain-path}
\end{figure}

\begin{figure}[!t]
\centering
\includegraphics[width=\columnwidth]{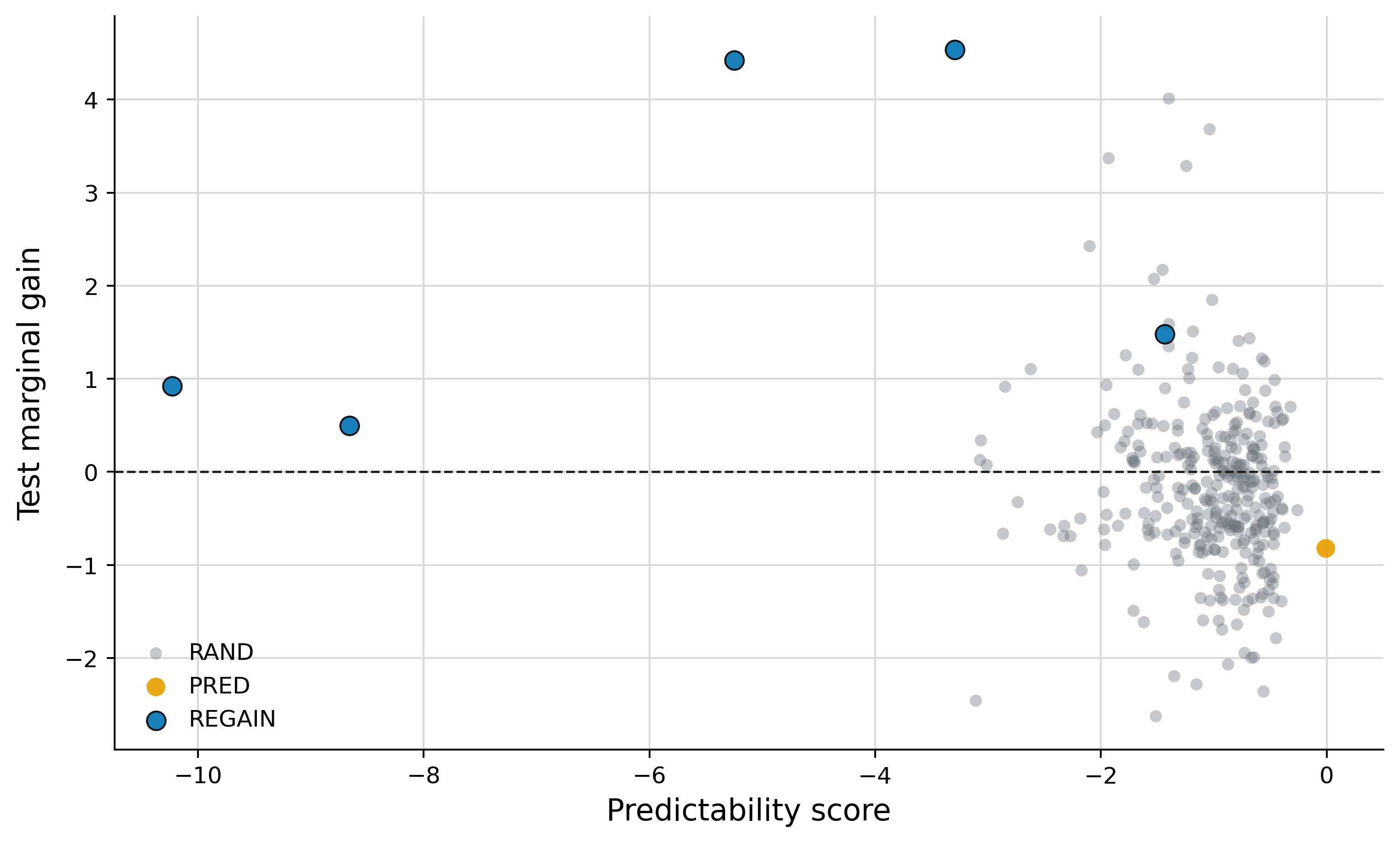}
\caption{Tourism predictability versus gain. Easier-to-forecast auxiliary series do not necessarily yield larger reconciliation gain.}
\label{fig:predictability-vs-gain}
\end{figure}

\begin{figure}[!t]
\centering
\includegraphics[width=\columnwidth]{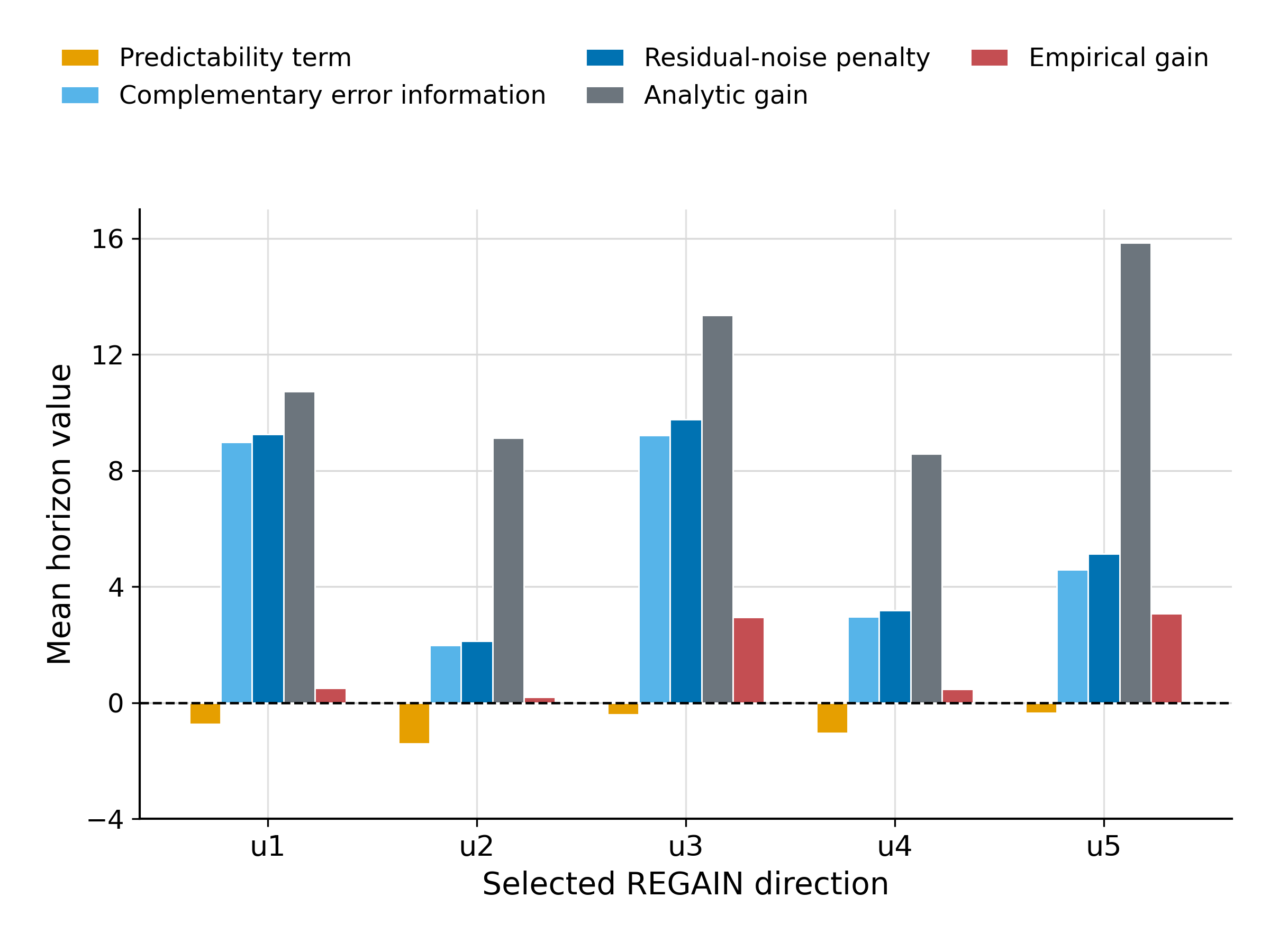}
\caption{Tourism gain decomposition for selected directions. Realized gain depends jointly on auxiliary predictability and complementary error information.}
\label{fig:gain-decomposition}
\end{figure}

\begin{figure}[!t]
\centering
\includegraphics[width=\columnwidth]{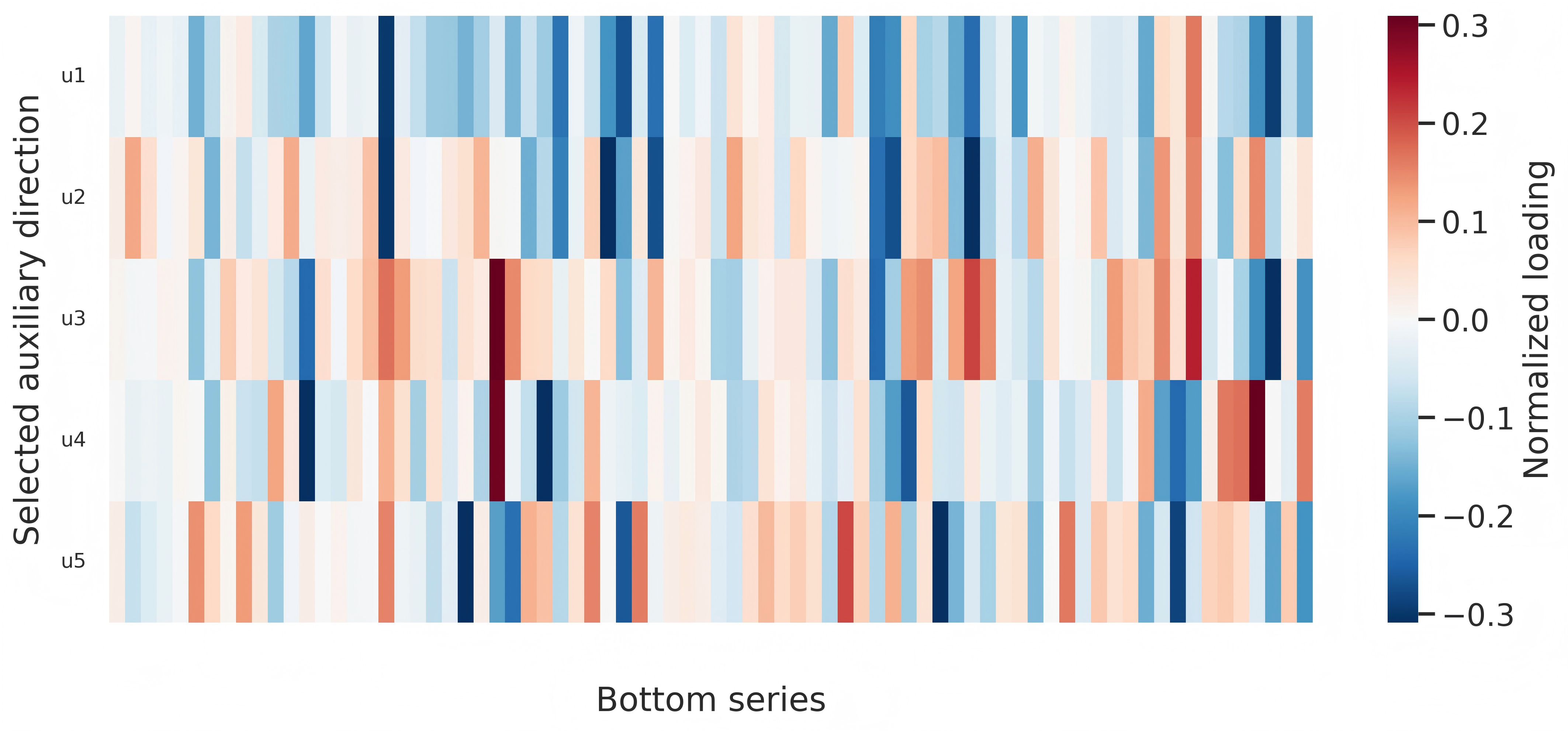}
\caption{Signed loading structure of the learned auxiliary directions on Tourism. Rows correspond to accepted auxiliary directions, columns correspond to bottom-level series, and colors indicate signed normalized loadings.}
\label{fig:direction-loadings}
\end{figure}

\begin{figure}[!t]
\centering
\includegraphics[width=\columnwidth]{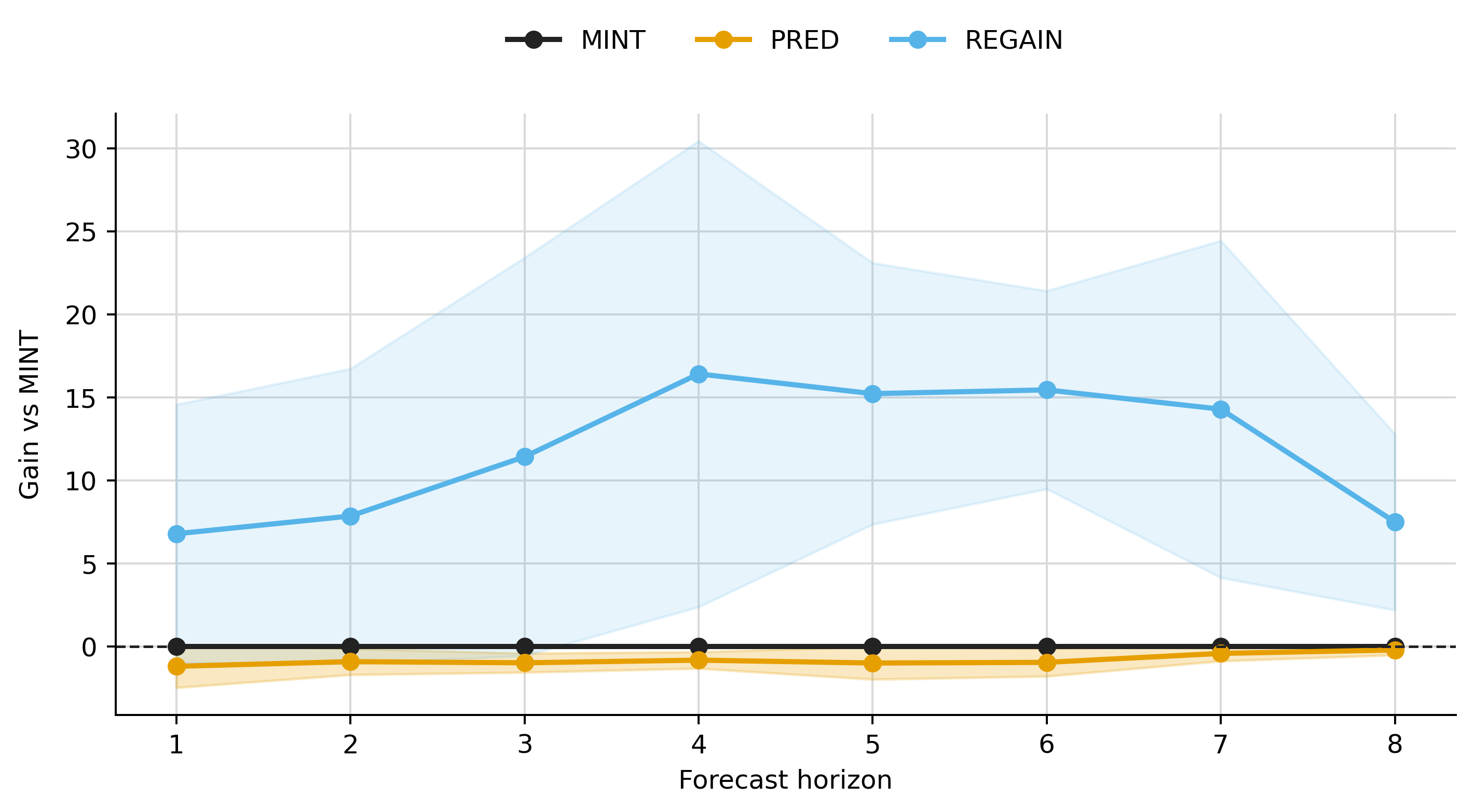}
\caption{Distribution of Tourism gains across horizons and target nodes. The figure shows how improvements vary across forecast horizons and target nodes.}
\label{fig:horizon-node-diagnostics}
\end{figure}


The diagnostic figures sharpen the mechanism behind the aggregate results.
Figures~\ref{fig:tourism-k-sweep} and~\ref{fig:marginal-gain-path} show that
the stagewise gains are front-loaded: the first few accepted directions
account for most of the improvement, while later additions mainly help on the
search split and contribute less reliably on selection and test. This is
consistent with the moderate-\(k\) operating region seen in
Figure~\ref{fig:tourism-k-sweep}, and helps explain why the optional joint
refinement adds only limited extra benefit once the early stagewise directions
already account for most of the observed gain.

Figure~\ref{fig:predictability-vs-gain} makes the main conceptual point visible in the Tourism data: standalone auxiliary predictability is at best an incomplete proxy for reconciliation usefulness. Some candidate directions are relatively easy to forecast but produce little realized gain, whereas the directions selected by \textsc{REGAIN} tend to lie closer to the upper envelope of gain at comparable predictability levels. Figure~\ref{fig:gain-decomposition} supports the same interpretation from a mechanistic side: directions become valuable when forecastability is combined with sufficiently strong complementary error information relative to the tourism-node residuals, rather than from predictability alone.

The learned directions also exhibit visible structure. Figure~\ref{fig:direction-loadings} shows signed patterns in the loading matrix rather than unstructured noise, which is useful for interpretability even though sparsity is not enforced. Finally, Figure~\ref{fig:horizon-node-diagnostics} shows that the gains are not confined to a single horizon or a tiny subset of nodes: improvement appears across multiple forecast horizons and is spread across the target series, supporting the view that the learned auxiliary measurements provide broadly useful information rather than a narrow local correction.
\FloatBarrier

\section{Discussion}

The results suggest that auxiliary-direction learning is best viewed as a
target-aware measurement-design layer for forecast reconciliation. Classical
reconciliation asks how to combine forecasts once the measurement system is
given; REGAIN asks which additional linear measurements are worth forecasting
because they reduce final target-node risk after reconciliation. This
distinction matters because high standalone predictability does not by itself
imply large reconciled gain. A useful auxiliary series must also expose
target-relevant residual uncertainty and carry error information that is not
redundant with the natural forecast block.

The experiments also clarify the roles of the two algorithmic components. The
stagewise procedure is the main estimator: it searches one direction at a time,
checks marginal gain on a held-out selection split, and stops when additional
directions no longer transfer reliably. The optional \textsc{REGAIN-JR}
refinement is better interpreted as a local post-processing step than as a
separate source of the method's value. Once the stagewise phase has found the
dominant directions, the remaining optimization headroom is small and more
exposed to surrogate mismatch, finite-sample covariance noise, and the local
geometry of the frozen oracle. This helps explain why joint refinement yields
modest, case-dependent improvements in the present experiments.

The learned directions should not be confused with a newly discovered hierarchy.
They are linear measurements chosen for their reconciliation effect, and their
usefulness does not require them to be nonnegative, sparse, or tree-structured.
This is an advantage for predictive accuracy, but it creates an interpretability
trade-off. If auxiliary measurements must correspond to named regions, business
units, or physically meaningful aggregates, sparsity, nonnegativity, grouping,
or hierarchy-like constraints can be added during search or post-processing,
possibly at the cost of downstream gain. The orthogonality constraint in REGAIN
should therefore be read as normalization and capacity control, not as an
assumption that useful measurements or their forecast errors are statistically
independent.

The framework is most likely to help when the natural forecasts leave
structured residual uncertainty that can be probed by additional forecastable
linear measurements. Gains may be small when the original measurement system
already captures the relevant uncertainty, when auxiliary series are too noisy,
or when candidate directions are redundant with the natural forecast errors. The
frozen-oracle design isolates the contribution of direction learning by scoring
all candidates under the same forecasting engine, but it also limits the scope
of the conclusions: different, jointly tuned, or end-to-end forecasting systems
could change both the learned directions and the realized gains.

Several limitations remain. The covariance-risk non-deterioration result is a
population covariance-risk statement under correctly specified GLS, not a
finite-sample MSE or test-loss guarantee. The empirical objective includes
bias changes,
whereas the cleanest analytical decomposition isolates covariance-side
mechanisms. The stability result relies on uniform perturbation assumptions for
the estimated covariance family, while the implementation uses finite rolling
splits, covariance shrinkage, validation screening, and ridge stabilization.
The current benchmarks provide evidence that gain-selected auxiliary directions
can help in both ordinary multivariate and hierarchical settings, but broader
regimes, constrained interpretable direction classes, and oracle-aware
refinements remain important directions for future work.

\section{Conclusion}

We introduced REGAIN, short for REconciliation GAIN-driven Auxiliary
Direction Learning, as a target-aware framework for learning auxiliary
measurements in forecast reconciliation. The central message is that an
auxiliary direction should be evaluated by the downstream reconciliation gain it
creates on the target nodes, not by standalone predictability alone. This
recasts auxiliary aggregation as a direction-learning problem: under a frozen
forecasting oracle, REGAIN searches over normalized directions and adds only
those whose marginal gain remains visible on held-out data.

The theoretical analysis supports this viewpoint by separating population
covariance-risk reduction, bias changes in realized quadratic risk, stability of
estimated covariance-gain signals, and validation-based screening of marginal
gain. The single-direction analysis further explains why useful measurements
must balance target exposure, complementary error information, and residual
noise. Empirically, the Beijing PM2.5 and Tourism studies show that
gain-selected directions can improve target-node accuracy in both ordinary
multivariate and hierarchical settings. Most of the observed improvement is
captured by the stagewise core, while joint refinement provides only limited and
case-dependent additional benefit. Overall, the results support the view that
forecast reconciliation can benefit not only from better projection rules under
a fixed structure, but also from learning which additional measurements should
enter the reconciliation system.

\bibliographystyle{plainnat}
\bibliography{refs}

\clearpage
\onecolumn
\appendix

\section{Proofs for the Gain Characterization and Optimization Properties}
\label{appendix:proof}

This appendix collects the detailed proofs omitted from the main text. The arguments are stated for the structured hierarchical or grouped setting, because this setting subsumes the ordinary multivariate case through an identity measurement structure. Concretely, the ordinary multivariate formulas are obtained by taking $S=I_n$, so that $y_t=b_t$, $\Meas_S(U)=[I_n;U^\top]$, and $W^S_h(U)$ is the joint covariance of the natural and auxiliary residuals.

\subsection{Main-text assumptions used in the proofs}

For a fixed horizon $h$, the proofs below use Assumptions~\ref{ass:wellposed} and~\ref{ass:uniform-stability} from the main text. We restate only the pieces used repeatedly in the algebra.

\begin{enumerate}
    \item Assumptions~\ref{ass:wellposed} imply that the residual covariances are well defined and the generalized least-squares projection problems are well posed. Concretely, $W_{yy,h}$ and $W^S_h(U)$ are symmetric positive definite, and the bottom-level normal matrices
    \[
    S^\top W_{yy,h}^{-1}S,
    \qquad
    \Meas_S(U)^\top (W^S_h(U))^{-1}\Meas_S(U)
    \]
    are nonsingular for the candidates under consideration.

    \item Corollary~\ref{cor:gain} uses the weighted covariance-risk functional
    \[
    \mathcal{R}^{\mathrm{cov}}_h(\varepsilon_h;Q_h)
    :=
    \tr\!\left(Q_h \, \mathrm{Cov}(\varepsilon_h)\right).
    \]
    In the structured proofs below, $Q_h$ acts on reported-node errors.
    The corresponding state-coordinate weight is denoted
    $Q^b_h=S^\top Q_hS$.
    More generally, for any square-integrable $\varepsilon_h$ with mean $\mu_h$,
    \[
    \mathbb{E}[\varepsilon_h^\top Q_h \varepsilon_h]
    =
    \tr\!\left(Q_h \, \mathrm{Cov}(\varepsilon_h)\right)
    + \mu_h^\top Q_h \mu_h.
    \]

    \item Proposition~\ref{prop:single} specializes to the single-direction case $k=1$. Then the auxiliary residual covariance block is scalar:
    \[
    W^S_h(u)=
    \begin{bmatrix}
    W_{yy,h} & k^S_h(u) \\
    k^S_h(u)^\top & r^S_h(u)
    \end{bmatrix},
    \]
    with $r^S_h(u)=\mathrm{Var}(e^{c,S}_{t,h}(u))$ and $k^S_h(u)=\mathrm{Cov}(e^y_{t,h},e^{c,S}_{t,h}(u))$. Its Schur complement is
    \[
    \tau^S_h(u)=r^S_h(u)-k^S_h(u)^\top W_{yy,h}^{-1}k^S_h(u)
    \]
    and $\tau^S_h(u)>0$ follows from $W^S_h(u)\succ 0$.
\end{enumerate}

\subsection{Proof of Lemma~\ref{lem:covariance}}

Write $\tilde b^{(0)}_{t,h}$ for the natural-only reconciled bottom-level estimate, so that $\tilde y^{(0)}_{t,h}=S\tilde b^{(0)}_{t,h}$. The natural-only measurement equation is
\[
\hat y_{t,h}=Sb_{t+h}-e^y_{t,h},
\qquad
\mathrm{Cov}(e^y_{t,h})=W_{yy,h}.
\]
The unregularized bottom-level GLS estimator is
\[
\tilde b^{(0)}_{t,h}
=
\big(S^\top W_{yy,h}^{-1}S\big)^{-1}
S^\top W_{yy,h}^{-1}\hat y_{t,h}.
\]
Subtracting $b_{t+h}$ yields
\[
\tilde b^{(0)}_{t,h}-b_{t+h}
=
-\big(S^\top W_{yy,h}^{-1}S\big)^{-1}
S^\top W_{yy,h}^{-1}e^y_{t,h}.
\]
Thus
\[
\mathrm{Cov}\!\big(\tilde b^{(0)}_{t,h}-b_{t+h}\big)
=
\big(S^\top W_{yy,h}^{-1}S\big)^{-1},
\]
which is the stated structured natural-only covariance $\Sigma^S_{0,h}$.

For the augmented structured system, stack the reported hierarchy and auxiliary residuals as
\[
e^{z,S}_{t,h}(U)=
\begin{bmatrix}
e^y_{t,h} \\
e^{c,S}_{t,h}(U)
\end{bmatrix},
\qquad
\mathrm{Cov}\!\big(e^{z,S}_{t,h}(U)\big)=W^S_h(U),
\]
so that
\[
\hat z^S_{t,h}(U)=\Meas_S(U)b_{t+h}-e^{z,S}_{t,h}(U).
\]
The unregularized bottom-level GLS estimator is
\[
\begin{aligned}
\tilde b_{t,h}(U)
&=
\bigl(\Meas_S(U)^\top (W^S_h(U))^{-1} \Meas_S(U)\bigr)^{-1} \\
&\quad {}\times \Meas_S(U)^\top (W^S_h(U))^{-1} \hat z^S_{t,h}(U).
\end{aligned}
\]
Subtracting $b_{t+h}$ and taking covariance gives
\[
\mathrm{Cov}\!\big(\tilde b_{t,h}(U)-b_{t+h}\big)
=
\big(\Meas_S(U)^\top (W^S_h(U))^{-1}\Meas_S(U)\big)^{-1}.
\]
This is the stated structured augmented covariance $\Sigma^S_{U,h}$.

With the identity measurement block $S=I_n$, the same calculation gives
the ordinary multivariate version with \(y_t=b_t\) and
\[
\Sigma^S_{0,h}=W_{yy,h},
\qquad
\Sigma^S_{U,h}
=
\big(\Meas_S(U)^\top (W^S_h(U))^{-1}\Meas_S(U)\big)^{-1}.
\]

\subsection{Proof of Corollary~\ref{cor:gain}}

Let
\[
\varepsilon^S_{0,t,h}=\tilde b^{(0)}_{t,h}-b_{t+h},
\qquad
\varepsilon^S_{U,t,h}=\tilde b_{t,h}(U)-b_{t+h}
\]
denote the bottom-level errors under the natural-only and augmented structured systems, respectively. By Lemma~\ref{lem:covariance},
\[
\mathrm{Cov}(\varepsilon^S_{0,t,h})=\Sigma^S_{0,h},
\qquad
\mathrm{Cov}(\varepsilon^S_{U,t,h})=\Sigma^S_{U,h}.
\]
The reported-node errors are
\[
\varepsilon^y_{0,t,h}=S\varepsilon^S_{0,t,h},
\qquad
\varepsilon^y_{U,t,h}=S\varepsilon^S_{U,t,h},
\]
so their covariances are
\[
\Sigma^y_{0,h}=S\Sigma^S_{0,h}S^\top,
\qquad
\Sigma^y_{U,h}=S\Sigma^S_{U,h}S^\top.
\]

Using the covariance-risk functional above,
\[
\mathcal{R}^{\mathrm{cov}}_{0,h}(Q_h)
\;=\;
\tr\!\left(Q_h \, \mathrm{Cov}(\varepsilon^y_{0,t,h})\right)
\;=\;
\tr\!\left(Q_h \Sigma^y_{0,h}\right),
\]
and likewise
\[
\mathcal{R}^{\mathrm{cov}}_{U,h}(Q_h)
\;=\;
\tr\!\left(Q_h \Sigma^y_{U,h}\right).
\]
The theoretical covariance-risk gain is therefore
\[
\mathcal G^{\mathrm{ana}}_{h,S}(U;Q_h)
=
\mathcal{R}^{\mathrm{cov}}_{0,h}(Q_h)-\mathcal{R}^{\mathrm{cov}}_{U,h}(Q_h).
\]
Substituting the two trace expressions gives
\[
\mathcal G^{\mathrm{ana}}_{h,S}(U;Q_h)
=
\tr\!\left(
Q_h(\Sigma^y_{0,h}-\Sigma^y_{U,h})
\right),
\]
equivalently
\[
\mathcal G^{\mathrm{ana}}_{h,S}(U;Q_h)
=
\tr\!\left(
Q^b_h(\Sigma^S_{0,h}-\Sigma^S_{U,h})
\right),
\qquad Q^b_h=S^\top Q_hS.
\]
This is the unified covariance-risk formula; the ordinary formula follows
by the same identity-block argument stated at the start of the appendix.

\subsection{Proof of Proposition~\ref{prop:psd}}

Write the block inverse of $W^S_h(U)$ as
\[
(W^S_h(U))^{-1}
=
\begin{bmatrix}
B^S_{11,h}(U)
&
B^S_{12,h}(U) \\
\big(B^S_{12,h}(U)\big)^\top
&
B^S_{22,h}(U)
\end{bmatrix},
\]
where
\[
T^S_h(U)
=R^S_h(U)-K^S_h(U)^\top W_{yy,h}^{-1}K^S_h(U),
\]
\[
B^S_{11,h}(U)
=W_{yy,h}^{-1}
+W_{yy,h}^{-1}K^S_h(U)T^S_h(U)^{-1}\times K^S_h(U)^\top W_{yy,h}^{-1},
\]
\[
B^S_{12,h}(U)
=-W_{yy,h}^{-1}K^S_h(U)T^S_h(U)^{-1},
\]
\[
B^S_{22,h}(U)
=T^S_h(U)^{-1}.
\]
By Assumption~\ref{ass:wellposed}, $W^S_h(U)\succ 0$, so its Schur complement $T^S_h(U)$ is also positive definite. Multiplying out $\Meas_S(U)^\top (W^S_h(U))^{-1}\Meas_S(U)$ gives
\[
\begin{aligned}
&\Meas_S(U)^\top (W^S_h(U))^{-1}\Meas_S(U)
= S^\top W_{yy,h}^{-1}S \\
&+ \bigl(U-S^\top W_{yy,h}^{-1}K^S_h(U)\bigr)T^S_h(U)^{-1} \times
\bigl(U-S^\top W_{yy,h}^{-1}K^S_h(U)\bigr)^\top .
\end{aligned}
\]
which is the structured identity with
\[
M^S_h(U)=U-S^\top W_{yy,h}^{-1}K^S_h(U).
\]
Since
\[
M^S_h(U)T^S_h(U)^{-1}M^S_h(U)^\top \succeq 0,
\]
we have
\[
\Meas_S(U)^\top (W^S_h(U))^{-1}\Meas_S(U)\succeq S^\top W_{yy,h}^{-1}S.
\]
The inverse reverses the Loewner order on positive definite matrices, so
\[
\Sigma^S_{U,h}
=
\big(\Meas_S(U)^\top (W^S_h(U))^{-1}\Meas_S(U)\big)^{-1}
\preceq
\big(S^\top W_{yy,h}^{-1}S\big)^{-1}
=
\Sigma^S_{0,h}.
\]
Hence
\[
\Sigma^S_{0,h}-\Sigma^S_{U,h}
\succeq 0.
\]
Finally, if $Q_h\succeq 0$, then $Q^b_h=S^\top Q_hS\succeq 0$ and
\[
\mathcal G^{\mathrm{ana}}_{h,S}(U;Q_h)
=
\tr\!\big(Q^b_h(\Sigma^S_{0,h}-\Sigma^S_{U,h})\big)\ge 0,
\]
because the trace of the product of two positive semidefinite matrices is nonnegative.

\subsection{Proof of Lemma~\ref{lem:bias-risk}}

Using the notation from the proof of Corollary~\ref{cor:gain}, set
\[
\mu^y_{0,h}=\E[\varepsilon^y_{0,t,h}],
\qquad
\mu^y_{U,h}=\E[\varepsilon^y_{U,t,h}].
\]
The identity
\[
\mathbb{E}[\varepsilon^\top Q_h \varepsilon]
=
\tr\!\left(Q_h\,\mathrm{Cov}(\varepsilon)\right)+\E[\varepsilon]^\top Q_h \E[\varepsilon]
\]
therefore gives
\[
\begin{aligned}
\E\!\left[\loss_{Q_h}\big(\tilde y^{(0)}_{t,h},y_{t+h}\big)\right]
&=
\tr\!\left(Q_h \, \mathrm{Cov}(\varepsilon^y_{0,t,h})\right)
+(\mu^y_{0,h})^\top Q_h \mu^y_{0,h}, \\
\E\!\left[\loss_{Q_h}\big(\tilde y_{t,h}(U),y_{t+h}\big)\right]
&=
\tr\!\left(Q_h \, \mathrm{Cov}(\varepsilon^y_{U,t,h})\right)
+(\mu^y_{U,h})^\top Q_h \mu^y_{U,h}.
\end{aligned}
\]
Substituting
\[
\mathrm{Cov}(\varepsilon^y_{0,t,h})=\Sigma^y_{0,h},
\qquad
\mathrm{Cov}(\varepsilon^y_{U,t,h})=\Sigma^y_{U,h},
\]
and subtracting the two displays yields
\[
\begin{aligned}
&\E\!\left[\loss_{Q_h}\big(\tilde y^{(0)}_{t,h},y_{t+h}\big)\right]
-\E\!\left[\loss_{Q_h}\big(\tilde y_{t,h}(U),y_{t+h}\big)\right] \\
&\qquad=
\tr\!\big(Q_h(\Sigma^y_{0,h}-\Sigma^y_{U,h})\big)
+(\mu^y_{0,h})^\top Q_h \mu^y_{0,h}
-(\mu^y_{U,h})^\top Q_h \mu^y_{U,h},
\end{aligned}
\]
which is the unified bias-risk decomposition.

\subsection{Proof of Lemma~\ref{lem:perturb}}

Fix a horizon $h$ and abbreviate the population and estimated information
matrices by
\[
\begin{aligned}
\mathcal I^S_{0,h}
&=S^\top W_{yy,h}^{-1}S,\\
\widehat{\mathcal I}^S_{0,h}
&=S^\top \widehat W_{yy,h}^{-1}S,
\end{aligned}
\]
and
\[
\begin{aligned}
\mathcal I^S_h(U)
&=
\Meas_S(U)^\top (W^S_h(U))^{-1}\Meas_S(U),\\
\widehat{\mathcal I}^S_h(U)
&=
\Meas_S(U)^\top
\big(\widehat W^S_h(U)\big)^{-1}
\Meas_S(U).
\end{aligned}
\]
By Assumption~\ref{ass:uniform-stability},
\[
\lambda_{\min}(W_{yy,h})\ge \underline\lambda_h,
\qquad
\inf_{U\in\mathcal U_k}
\lambda_{\min}\!\big(W^S_h(U)\big)\ge \underline\lambda_h .
\]
For \(\delta_h<\underline\lambda_h\), Weyl's inequality implies
\[
\begin{aligned}
\lambda_{\min}(\widehat W_{yy,h})
&\ge \underline\lambda_h-\delta_h,\\
\inf_{U\in\mathcal U_k}
\lambda_{\min}\!\big(\widehat W^S_h(U)\big)
&\ge \underline\lambda_h-\delta_h.
\end{aligned}
\]
Thus the estimated covariance matrices are positive definite in this
regime. Set
\[
\rho_h(\delta_h)
=
\frac{\delta_h}{\underline\lambda_h(\underline\lambda_h-\delta_h)} .
\]
The resolvent identity gives
\[
\|\widehat W_{yy,h}^{-1}-W_{yy,h}^{-1}\|_{\mathrm{op}}
\le
\rho_h(\delta_h)
\]
and, uniformly over \(U\in\mathcal U_k\),
\[
\begin{aligned}
&\big(\widehat W^S_h(U)\big)^{-1}
-
\big(W^S_h(U)\big)^{-1}\\
&\quad =
\big(\widehat W^S_h(U)\big)^{-1}
\big(W^S_h(U)-\widehat W^S_h(U)\big)
\big(W^S_h(U)\big)^{-1},
\end{aligned}
\]
hence
\[
\sup_{U\in\mathcal U_k}
\left\|
\big(\widehat W^S_h(U)\big)^{-1}
-
\big(W^S_h(U)\big)^{-1}
\right\|_{\mathrm{op}}
\le
\rho_h(\delta_h).
\]
Using the fixed bound on \(\|S\|_{\mathrm{op}}\) and the uniform bound on
\(\|\Meas_S(U)\|_{\mathrm{op}}\),
\[
\begin{aligned}
\|\widehat{\mathcal I}^S_{0,h}-\mathcal I^S_{0,h}\|_{\mathrm{op}}
&\le
\|S\|_{\mathrm{op}}^2\rho_h(\delta_h),\\
\sup_{U\in\mathcal U_k}
\|\widehat{\mathcal I}^S_h(U)-\mathcal I^S_h(U)\|_{\mathrm{op}}
&\le
B_{\Meas,h}^2\rho_h(\delta_h).
\end{aligned}
\]
After restricting, if necessary, to
\(\delta_h\le \underline\lambda_h/2\), both information-matrix
perturbations are bounded by \(\tilde C_h\delta_h\) for a finite constant
\(\tilde C_h\).

By Proposition~\ref{prop:psd},
\[
\begin{aligned}
\mathcal I^S_h(U)
&=
\mathcal I^S_{0,h}
+M^S_h(U)(T^S_h(U))^{-1}M^S_h(U)^\top\\
&\succeq \mathcal I^S_{0,h}.
\end{aligned}
\]
Since \(\mathcal I^S_{0,h}\) is positive definite by
Assumption~\ref{ass:wellposed}, define
\[
\underline\iota_h=\lambda_{\min}(\mathcal I^S_{0,h})>0.
\]
Then
\[
\inf_{U\in\mathcal U_k}\lambda_{\min}(\mathcal I^S_h(U))
\ge \underline\iota_h.
\]
Choose \(c_h>0\) such that
\[
c_h\le \frac{\underline\lambda_h}{2},
\qquad
\tilde C_hc_h\le \frac{\underline\iota_h}{2}.
\]
For \(\delta_h\le c_h\), both \(\widehat{\mathcal I}^S_{0,h}\) and
\(\widehat{\mathcal I}^S_h(U)\) are positive definite, uniformly over
\(U\in\mathcal U_k\). Thus the estimated state covariance matrices in the
main text are well defined. Applying the resolvent identity again gives
\[
\widehat\Sigma^S_{0,h}-\Sigma^S_{0,h}
=
\big(\widehat{\mathcal I}^S_{0,h}\big)^{-1}
\big(\mathcal I^S_{0,h}-\widehat{\mathcal I}^S_{0,h}\big)
\big(\mathcal I^S_{0,h}\big)^{-1}
\]
and
\[
\begin{aligned}
\widehat\Sigma^S_{U,h}-\Sigma^S_{U,h}
&=
\big(\widehat{\mathcal I}^S_h(U)\big)^{-1}\\
&\quad {}\times
\big(\mathcal I^S_h(U)-\widehat{\mathcal I}^S_h(U)\big)
\big(\mathcal I^S_h(U)\big)^{-1}.
\end{aligned}
\]
Therefore
\[
\begin{aligned}
\|\widehat\Sigma^S_{0,h}-\Sigma^S_{0,h}\|_{\mathrm{op}}
&+
\sup_{U\in\mathcal U_k}
\|\widehat\Sigma^S_{U,h}-\Sigma^S_{U,h}\|_{\mathrm{op}}\\
&\le
\frac{2\tilde C_h}
{\underline\iota_h(\underline\iota_h-\tilde C_h\delta_h)}
\delta_h\\
&\le
\frac{4\tilde C_h}{\underline\iota_h^2}\delta_h\\
&=:
C_{\Sigma,h}\delta_h,
\end{aligned}
\]
which proves the covariance perturbation claim.

For the gain claim, write
\[
G_h(U)=\mathcal G^{\mathrm{ana}}_{h,S}(U;Q_h),
\qquad
\widehat G_h(U)=\widehat{\mathcal G}^{\mathrm{ana}}_{h,S}(U;Q_h).
\]
Using \(Q^b_h=S^\top Q_hS\),
\[
\begin{aligned}
\widehat G_h(U)-G_h(U)
&=
\tr\!\left(
Q^b_h\left[
(\widehat\Sigma^S_{0,h}-\Sigma^S_{0,h})
-
(\widehat\Sigma^S_{U,h}-\Sigma^S_{U,h})
\right]
\right).
\end{aligned}
\]
The trace inequality
\(|\tr(AB)|\le \mathrm{rank}(A)\|A\|_{\mathrm{op}}\|B\|_{\mathrm{op}}\),
together with the covariance perturbation claim, gives
\[
\sup_{U\in\mathcal U_k}
\left|\widehat G_h(U)-G_h(U)\right|
\le
\mathrm{rank}(Q^b_h)\|Q^b_h\|_{\mathrm{op}}\,
C_{\Sigma,h}\delta_h.
\]
Absorbing the fixed factor into a new constant yields
\[
\sup_{U\in\mathcal U_k}
\left|\widehat G_h(U)-G_h(U)\right|
\le
C_{G,h}\delta_h,
\]
as claimed.

\subsection{Proof of Corollary~\ref{cor:uniform-gain}}

By definition,
\[
\widehat G(U)-G(U)
=
\sum_{h=1}^H \omega_h\big(\widehat G_h(U)-G_h(U)\big).
\]
Taking absolute values, using $\omega_h\ge 0$, and applying Lemma~\ref{lem:perturb},
\[
\begin{aligned}
\sup_{U\in\mathcal U_k}
\left|\widehat G(U)-G(U)\right|
&\le
\sum_{h=1}^H \omega_h
\sup_{U\in\mathcal U_k}
\left|\widehat G_h(U)-G_h(U)\right| \\
&\le
\sum_{h=1}^H \omega_h C_{G,h}\delta_h \\
&\le
C\sum_{h=1}^H \omega_h\delta_h,
\end{aligned}
\]
where $C=\max_h C_{G,h}$.

\subsection{Proof of Proposition~\ref{prop:single}}

Now specialize to $k=1$ and write
\[
\Meas_S(u)=
\begin{bmatrix}
S \\
u^\top
\end{bmatrix},
\qquad
W^S_h(u)=
\begin{bmatrix}
W_{yy,h} & k^S_h(u) \\
k^S_h(u)^\top & r^S_h(u)
\end{bmatrix}.
\]
For this proof, write \(\Sigma^S_{u,h}\) for the augmented covariance
\(\Sigma^S_{U,h}\) evaluated at \(U=u\). Under
Assumption~\ref{ass:wellposed}, the Schur complement
\[
\tau^S_h(u)
=
r^S_h(u)-k^S_h(u)^\top W_{yy,h}^{-1}k^S_h(u)
\]
is strictly positive because \(W^S_h(u)\) is positive definite. Define the
upper-left block of \((W^S_h(u))^{-1}\) by
\[
B^S_{11,h}(u)
=
W_{yy,h}^{-1}
+W_{yy,h}^{-1}k^S_h(u)
\big(\tau^S_h(u)\big)^{-1}
k^S_h(u)^\top W_{yy,h}^{-1}.
\]
The block inverse formula gives
\[
(W^S_h(u))^{-1}
=
\begin{bmatrix}
B^S_{11,h}(u)
&
-W_{yy,h}^{-1}k^S_h(u)\big(\tau^S_h(u)\big)^{-1} \\
-\big(\tau^S_h(u)\big)^{-1}k^S_h(u)^\top W_{yy,h}^{-1}
&
\big(\tau^S_h(u)\big)^{-1}
\end{bmatrix}.
\]

Introduce the shorthand
\[
\tilde u^S_h=u-S^\top W_{yy,h}^{-1}k^S_h(u),
\qquad
\Sigma^S_{0,h}=\big(S^\top W_{yy,h}^{-1}S\big)^{-1}.
\]
Multiplying out the single-direction information matrix and grouping terms
yields
\[
\begin{aligned}
\mathcal I^S_h(u)
&:=
\Meas_S(u)^\top (W^S_h(u))^{-1}\Meas_S(u)\\
&=
S^\top W_{yy,h}^{-1}S
+\big(\tau^S_h(u)\big)^{-1}
\tilde u^S_h(\tilde u^S_h)^\top\\
&=
\big(\Sigma^S_{0,h}\big)^{-1}
+\big(\tau^S_h(u)\big)^{-1}
\tilde u^S_h(\tilde u^S_h)^\top .
\end{aligned}
\]
Hence, by Lemma~\ref{lem:covariance},
\[
\Sigma^S_{u,h}
=
\big(\mathcal I^S_h(u)\big)^{-1}
=
\Big(
\big(\Sigma^S_{0,h}\big)^{-1}
+\big(\tau^S_h(u)\big)^{-1}
\tilde u^S_h(\tilde u^S_h)^\top
\Big)^{-1}.
\]

Applying the Woodbury identity to this rank-one update gives
\[
\Sigma^S_{u,h}
=
\Sigma^S_{0,h}
-\frac{
\Sigma^S_{0,h}\tilde u^S_h(\tilde u^S_h)^\top \Sigma^S_{0,h}
}{
\tau^S_h(u)+(\tilde u^S_h)^\top \Sigma^S_{0,h}\tilde u^S_h
}.
\]
Set \(x^S_h(u)=\Sigma^S_{0,h}\tilde u^S_h\). Rearranging,
\[
\Sigma^S_{0,h}-\Sigma^S_{u,h}
=
\frac{
x^S_h(u)\big(x^S_h(u)\big)^\top
}{
\tau^S_h(u)+(\tilde u^S_h)^\top x^S_h(u)
}.
\]

Substituting this identity into the covariance-risk reduction formula from
Corollary~\ref{cor:gain}, with \(Q^b_h=S^\top Q_hS\), gives
\[
\mathcal G^{\mathrm{ana}}_{h,S}(u;Q_h)
=
\tr\!\left(
Q^b_h(\Sigma^S_{0,h}-\Sigma^S_{u,h})
\right).
\]
Therefore,
\[
\mathcal G^{\mathrm{ana}}_{h,S}(u;Q_h)
=
\frac{
\tr\!\left(
Q^b_h x^S_h(u)\big(x^S_h(u)\big)^\top
\right)
}{
\tau^S_h(u)+(\tilde u^S_h)^\top x^S_h(u)
}.
\]
Using the cyclic property of the trace and the identity
\(\tr(Axx^\top)=x^\top A x\) for conformable matrices, we obtain
\[
\mathcal G^{\mathrm{ana}}_{h,S}(u;Q_h)
=
\frac{
\big(x^S_h(u)\big)^\top Q^b_h x^S_h(u)
}{
\tau^S_h(u)+(\tilde u^S_h)^\top x^S_h(u)
},
\]
which is the structured single-direction formula.

The denominator is positive because \(\tau^S_h(u)>0\) and
\(\Sigma^S_{0,h}\succ0\). The expression also makes the qualitative
interpretation in the main text precise: the denominator combines the
effective residual-noise penalty with the baseline uncertainty seen along
the adjusted direction, whereas the numerator measures the adjusted
direction's weighted unresolved uncertainty in state coordinates.

\subsection{Proof of Proposition~\ref{prop:validation}}

Consider the event
\[
\mathcal E_j=
\left\{
\sup_{v\in\mathcal C_j}
\left|
\widehat\Delta^{\mathrm{sel}}_j(v)
-
\Delta^{\mathrm{pop}}_j(v)
\right|
\le \varepsilon_{j,n}
\right\},
\]
which satisfies \(\Pr(\mathcal E_j)\ge 1-\alpha_j\) by assumption. On
\(\mathcal E_j\), if a candidate \(v_j^\star\) is accepted with
\[
\widehat\Delta^{\mathrm{sel}}_j(v_j^\star)>\tau_j,
\qquad
\tau_j>\varepsilon_{j,n},
\]
then
\[
\Delta^{\mathrm{pop}}_j(v_j^\star)
\ge
\widehat\Delta^{\mathrm{sel}}_j(v_j^\star)-\varepsilon_{j,n}
>
\tau_j-\varepsilon_{j,n}
>
0.
\]
Hence acceptance implies
\(\Delta^{\mathrm{pop}}_j(v_j^\star)>0\) on the event
\(\mathcal E_j\). Therefore a false acceptance can occur only on
$\mathcal E_j^c$, and
\[
\Pr\!\left(
\Delta^{\mathrm{pop}}_j(v_j^\star)\le 0
\;\text{and}\;
v_j^\star\ \text{is accepted}
\right)
\le \alpha_j,
\]
which is the desired false-acceptance control.

\subsection{Well-posedness of the normalized Stiefel problem}
\label{app:stiefel-wellposedness}

Assume \(1 \le k \le n\), \(D\succ 0\), and that the map
\(V\mapsto \Gain(D^{-1/2}V)\) is continuous on \(\St(n,k)\).
Then \(\St(n,k)\) is nonempty; for example, the matrix formed by the
first \(k\) columns of the \(n\times n\) identity belongs to \(\St(n,k)\).

By definition,
\[
\St(n,k)=\{V \in \R^{n \times k}: V^\top V = I_k\}.
\]
The constraint map $V \mapsto V^\top V$ is continuous, and $\{I_k\}$ is closed in $\R^{k \times k}$, so $\St(n,k)$ is closed in $\R^{n \times k}$. In addition, every $V \in \St(n,k)$ satisfies
\[
\|V\|_F^2=\tr(V^\top V)=\tr(I_k)=k,
\]
hence $\St(n,k)$ is bounded. By Heine--Borel, $\St(n,k)$ is compact as a subset of the Euclidean space $\R^{n \times k}$.

Now consider the objective
\[
f(V)=\Gain(D^{-1/2}V).
\]
By the stated continuity condition, \(f\) is continuous on the nonempty
compact set \(\St(n,k)\). The extreme-value theorem therefore implies that there exists some $V^\star \in \St(n,k)$ such that
\[
f(V^\star)=\max_{V \in \St(n,k)} f(V).
\]
Equivalently, Problem~\ref{prob:normalized-gain} admits at least one maximizer.

\section{Experimental and implementation details}
\label{app:implementation}

This appendix collects the implementation details needed to reproduce the experiments. We keep these details separate from the theoretical derivations.

\subsection{Dataset Description and Preprocessing}
\label{app:data_preprocessing}

\textbf{Beijing PM2.5}
We use the Beijing Multi-Site Air Quality dataset from the UCI Machine
Learning Repository\footnote{\url{https://archive.ics.uci.edu/dataset/501/beijing+multi+site+air+quality+data}} and retain the PM2.5 channel from the 12 monitoring stations Aotizhongxin, Changping, Dingling, Dongsi, Guanyuan, Gucheng, Huairou, Nongzhanguan, Shunyi, Tiantan, Wanliu, and Wanshouxigong. The raw hourly observations are aggregated to daily frequency by averaging all available hourly PM2.5 values within each station-day. This gives a wide daily table with 1461 days, from 2013-03-01 to 2017-02-28, and 12 observed station series. Days with no valid hourly PM2.5 value for a station are left missing after daily aggregation and are imputed at loading time by forward fill, backward fill, and finally zero fill if needed. Before imputation, there are 54 missing station-days out of $(1461\times 12)$ station-days, accounting for about 0.31\% of the daily station-level observations.

\textbf{Tourism}
For Tourism, we use the Australian domestic tourism data derived from
Rob J Hyndman's public data repository\footnote{\url{https://robjhyndman.com/data/TourismData_v4.csv}}.
The data measure Australian domestic tourism flows, commonly interpreted
as the number of overnight trips by Australian residents away from home.
Following the standard geographic hierarchy used in forecast-reconciliation
studies, the national total is disaggregated into states and territories,
then into tourism zones, and finally into regional bottom-level series.
The monthly sample spans 1998-01 to 2017-12, giving 240 observations for
each series. The processed hierarchy contains 76 bottom-level regional series and 35 upper-level aggregate series.
We keep this hierarchy as the natural reconciliation structure and learn
auxiliary directions only as additional measurements outside the original
geographic hierarchy.

\subsection{Frozen Forecasting Oracles}
\label{app:frozen_oracles}
All oracle calls are made without fine-tuning. The same cached base forecasts
are shared by all methods within each oracle and forecast-mode setting.

Chronos2 uses the Hugging Face checkpoint \path{amazon/chronos-2}, implemented with \texttt{chronos-forecasting==2.2.2} and \texttt{float32}. We evaluate both
independent and grouped modes. In independent mode, each channel is treated as a univariate series; in grouped mode, all channels from the same rolling-origin
window share the same group id. The median quantile is used as the point
forecast.

 TimesFM uses the 2.5-200M Transformers checkpoint with the Transformers backend
and \texttt{float32}; the checkpoint identifier is \path{google/timesfm-2.5-200m-transformers}. We use it as a frozen univariate oracle by flattening the multivariate batch into independent series. Contexts shorter than the checkpoint patch length 32 are left-padded. No TimesFM joint multivariate mode is used.

\subsection{Baseline Implementations}
\label{app:baselines}

The base forecasts are the frozen-oracle forecasts before any reconciliation or auxiliary augmentation. For Tourism, the oracle is applied directly to all series in the supplied hierarchy, so the resulting base forecasts may be incoherent across aggregation levels. For PM2.5, the base forecasts are the oracle forecasts for the 12 station-level series.

MINT is implemented as MINT-shrinkage when a nontrivial hierarchy is
available. For Tourism, the forecast-error covariance is estimated on
\(\Tfit\) using the same Ledoit--Wolf shrinkage covariance estimator as in REGAIN, and use the standard MINT reconciliation map to project the base forecasts onto the hierarchy.

For RAND, FLAP, and PRED, \(k\in\{1,2,4,6,8\}\) denotes the fixed number
of auxiliary series added to the reconciliation system. 

Random directions are sampled in the same whitened coordinate system as REGAIN. Specifically, we sample \(V\in\mathrm{St}(n,k)\) by QR-orthogonalizing a Gaussian matrix and map it back to the original bottom-series coordinates as \(U=D^{-1/2}V\). Random directions therefore use the same whitening and normalization as REGAIN. For each \(k\in\{1,2,4,6,8\}\), three restarts are evaluated and the best search-split restart is carried to selection and test evaluation.

FLAP uses principal-component directions estimated only from
\(\Tfit\), never from the selection or test splits. For Beijing PM2.5, PCA is applied directly to the
original station-series matrix. For Tourism, PCA is applied to the full
hierarchy matrix \(Y_{\rm fit}=B_{\rm fit}S^\top\), including both upper and
bottom series.
Writing the first \(k\) PCA
loadings as \(\Phi_k=[\phi_1,\ldots,\phi_k]\in\mathbb R^{N\times k}\), the corresponding bottom-level auxiliary directions are $U_{\rm FLAP}=S^\top \Phi_k$. The resulting directions are normalized using the same convention as the other auxiliary-direction baselines.

The predictability baseline uses the same candidate-generation machinery as
REGAIN, but replaces the direct reconciliation-gain objective by auxiliary forecastability. For a candidate \(U\), it scores the auxiliary forecasts by
\[
-\frac{1}{|\mathcal A|H}\sum_{t\in\mathcal A}\sum_{h=1}^{H}
\left\|c_{t+h}(U)-\hat c_{t,h}(U)\right\|_2^2 .
\]
After the directions are selected, the final reconciliation evaluation uses the same covariance estimation and augmented reconciliation procedure as the other
auxiliary-direction methods.

\newpage
\subsection{Random Seeds and Computing Environment}
\label{app:seeds_environment}

Unless otherwise noted, all main multi-seed experiments were conducted using the
random seeds
\[
\{39,40,41,42,43,44\}.
\]
For reproducibility, experiments were run with Python 3.11, PyTorch
2.8.0+cu128, NumPy 2.4.3, pandas 3.0.2, \texttt{chronos-forecasting} 2.2.2,
Transformers 4.57.6, Accelerate 1.13.0, Hugging Face Hub 0.36.2,
Safetensors 0.7.0, and TimesFM 2.0.0. All experiments were executed on NVIDIA
Tesla V100-SXM2-16GB GPUs with driver version 580.82.07.

\subsection{Runtime Summary}
\label{app:runtime_summary}

Table~\ref{tab:runtime-summary} reports the wall-clock runtime of REGAIN and
REGAIN-JR under the available oracle settings. Runtime is measured in minutes
under the computing environment described in Appendix~\ref{app:seeds_environment}.

\begin{table}[!htbp]
\centering
\caption{Runtime summary of REGAIN and REGAIN-JR. Runtime is reported in minutes as \textsc{REGAIN}/\textsc{REGAIN-JR}.}
\label{tab:runtime-summary}
\small
\setlength{\tabcolsep}{6pt}
\begin{tabular}{clc}
\toprule
Dataset & Setting & Runtime (mins) \\
\midrule
\multirow{3}{*}{Beijing PM2.5}
& Chronos2 indep. & 164.2 / 177.0 \\
& Chronos2 joint  & 204.9 / 240.8 \\
& TimesFM indep.  & 422.2 / 459.7 \\
\midrule
\multirow{3}{*}{Tourism}
& Chronos2 indep. & 5.3 / 5.8 \\
& Chronos2 joint  & 6.3 / 7.2 \\
& TimesFM indep.  & 6.2 / 7.6 \\
\bottomrule
\end{tabular}
\end{table}

\end{document}